\documentclass[journal]{IEEEtran}
\usepackage{amsmath,amsfonts}
\usepackage{algorithmic}
\usepackage{algorithm}
\usepackage{array}
\usepackage[caption=false,font=normalsize,labelfont=sf,textfont=sf]{subfig}
\usepackage{textcomp}
\usepackage{stfloats}
\usepackage{url}
\usepackage{verbatim}
\usepackage{graphicx}
\usepackage{cite}

\usepackage{amsmath}
\usepackage{amssymb}
\usepackage{booktabs}
\usepackage{textcomp}
\usepackage{multirow}
\usepackage{gensymb}
\usepackage{ulem}
\usepackage{makecell}
\usepackage{enumerate}
\usepackage{bbm}
\usepackage{pifont}
\usepackage{xcolor}
\usepackage[numbers]{natbib}
\usepackage{hyperref}
\hypersetup{
  colorlinks   = true, 
  urlcolor     = blue, 
  linkcolor    = red, 
  citecolor   = red 
}

\newcommand{\eg}{\textit{e}.\textit{g}.}

\begin{document}

\title{Close the {Optical Sensing} Domain Gap by Physics-Grounded Active Stereo Sensor Simulation}

\author{Xiaoshuai Zhang*$^{1}$, Rui Chen*$^{2}$, Ang Li**$^{1}$, Fanbo Xiang**$^{1}$, Yuzhe Qin**$^{1}$, Jiayuan Gu**$^{1}$, Zhan Ling**$^{1}$, Minghua Liu**$^{1}$, Peiyu Zeng**$^{2}$, Songfang Han***$^{1}$, Zhiao Huang***$^{1}$, Tongzhou Mu***$^{1}$, Jing Xu$^{2}$, and Hao Su$^{1}$
\thanks{$^{1}$Xiaoshuai Zhang, Ang Li, Fanbo Xiang, Yuzhe Qin, Jiayuan Gu, Zhan Ling, Minghua Liu, Songfang Han, Zhiao Huang, Tongzhou Mu, Hao Su are with Department of Computer Science and Engineering, University of California, San Diego, CA, USA}
\thanks{$^{2}$Rui Chen, Peiyu Zeng, Jing Xu are with Department of Mechanical Engineering, Tsinghua University, Beijing, China}
\thanks{Corresponding authors: Jing Xu, Hao Su.}
\thanks{* These authors contributed equally.}
\thanks{** These authors contributed equally.}
\thanks{*** These authors contributed equally.}
\thanks{Manuscript received July 5, 2022; revised November 16, 2022.}
}



\maketitle

\begin{abstract}
In this paper, we focus on the simulation of active stereovision depth sensors, which are popular in both academic and industry communities. Inspired by the underlying mechanism of the sensors, we designed a fully physics-grounded simulation pipeline that includes material acquisition, ray-tracing-based infrared (IR) image rendering, IR noise simulation, and depth estimation. 

The pipeline is able to generate depth maps with material-dependent error patterns similar to a real depth sensor in real time.
We conduct real experiments to show that perception algorithms and reinforcement learning policies trained in our simulation platform could transfer well to the real-world test cases without any fine-tuning. Furthermore, due to the high degree of realism of this simulation, our depth sensor simulator can be used as a convenient testbed to evaluate the algorithm performance in the real world, which will largely reduce the human effort in developing robotic algorithms. The entire pipeline has been integrated into the SAPIEN simulator and is open-sourced to promote the research of vision and robotics communities{\footnote{{Kuafu Real-Time Ray Tracing Renderer: \href{https://github.com/jetd1/kuafu}{https://github.com/jetd1/kuafu} \\ SimSense depth computing module: \href{https://github.com/angli66/simsense}{https://github.com/angli66/simsense} \\ SAPIEN simulator: \href{https://github.com/haosulab/SAPIEN}{https://github.com/haosulab/SAPIEN}}}}. 
\end{abstract}


\section{Introduction}
\IEEEPARstart{R}{ecently}, researchers in the robotics community are becoming increasingly interested in training robots in simulation environments, due to the convenience of conducting large-scale experiments (\textit{e.g.} grasp label generation~{\cite{miller2004graspit, eppner2019billion, qin2020s4g,fang2020graspnet,8972562,newbury2022deep}} and object manipulation by deep reinforcement learning (DRL)~\cite{doi:10.1177/0278364919872545, chen2021a, mu2021maniskill}). 

\begin{figure}[htbp]
    \centering
    \includegraphics[width=0.9\linewidth]{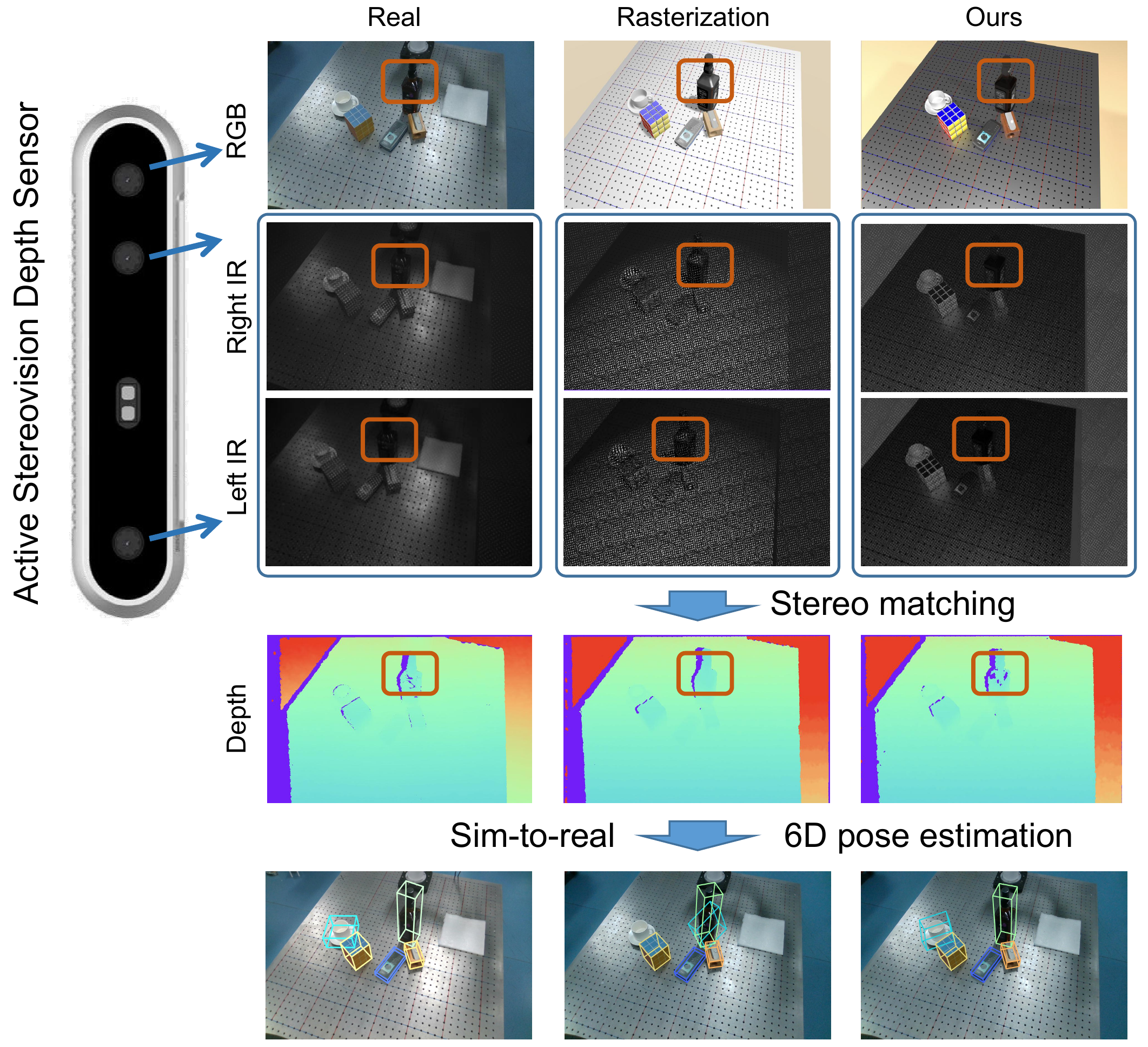}
    \caption{The depth measurement of real active stereovision depth sensors is greatly affected by the object materials. The projected pattern is not well reflected on the translucent or transparent surface, resulting in incomplete measurement. However, conventional rasterization rendering cannot simulate {this} phenomenon. In this paper, we aim to close the optical sensing domain gap by building an active stereovision depth sensor simulator in a physics-grounded fashion, so that it can simulate the material-dependent depth errors more accurately and improve the sim-to-real performance of 6D pose estimation algorithms. \textbf{(Zoom in to better see the IR patterns and pose estimation results)}}
    \label{fig:teaser}
\end{figure}

In this work, we aim to achieve low {optical sensing} domain gap (not physical domain gap) of robot simulators with a light-weight pipeline. We drive the exploration by important applications: 1) Given CAD models of objects, the robot simulator generates a large amount of simulated data, on which learning-based perception models (\textit{e.g.} object detection, pose estimation) and policy learning algorithms (\textit{e.g.} object grasping) are trained and can work in the real world without any fine-tuning; 2) Before carrying out real-world experiments, researchers can evaluate perception or policy learning algorithms in simulation, which requires that the performance in the real world and the simulator should be consistent. 
\begin{figure*}[htbp]
    \centering
    \includegraphics[width=\textwidth]{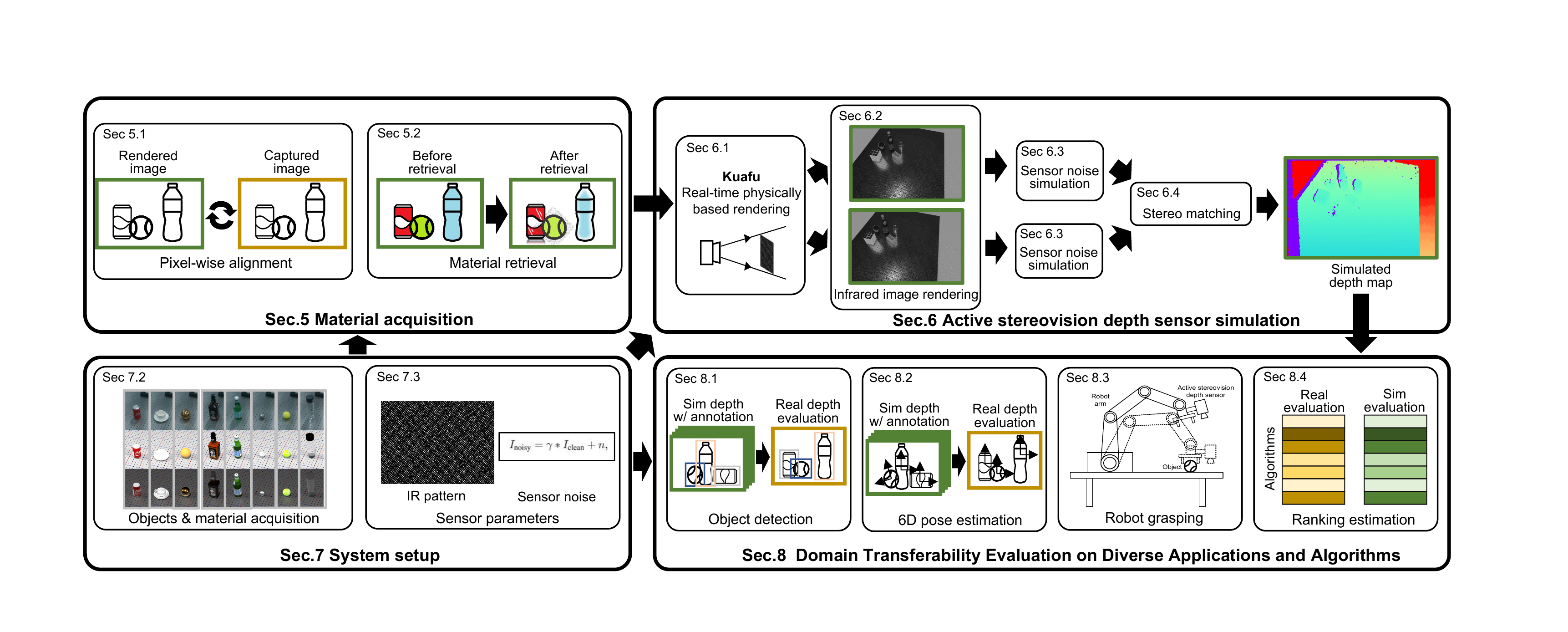}
    \caption{Paper roadmap. In this paper, we study how to build an active stereovision depth sensor simulator with a low optical sensing domain gap. In Section~\ref{sec:material_acquisition}, we describe the material acquisition method, which includes pixel-wise alignment between the simulation and the real world and PBR material retrieval based on a novel multi-spectral loss. In Section~\ref{sec:sensor_simulation}, we establish an active stereovision depth sensor simulator by following the same mechanism of real sensors, which includes IR pattern projection, transport of IR lights by ray tracing, IR camera sensor noise and depth generation by stereo matching. Our new customizable renderer \textit{Kuafu} is covered in that section. In Section~\ref{sec:system_setup}, we describe the real-world and simulated system setup and the preparation for experiments. In Section~\ref{sec:applications}, we conduct experiments of four applications of our depth sensor simulator.}
    \label{fig:paper_structure}
\end{figure*}
We choose to narrow the sim-to-real optical sensing domain gap by studying the simulation of depth sensors, more precisely, active stereovision depth sensors. This choice distinguishes us from most recent efforts except for few ones. We list five reasons to simulate active stereovision depth sensors instead of RGB sensors (cameras) or other depth sensors. The first two reasons justify the choice of depth sensor over RGB camera, and the last three justify the choice of active stereovision sensor over other depth sensors: \\
1) Depth map is more suitable for robotic tasks. For many robotic tasks that require interaction, 3D geometry information of the physical world is of utmost importance~\cite{schmidt2015depth, qin2020s4g, doi:10.1177/0278364919859066, cong2021comprehensive}. However, this information cannot be retrieved easily from RGB cameras;\\
2) Depth maps do not involve color information, which is extremely challenging to align between the simulator and the real world.
A few recent papers \cite{viereck2017learning, 8794226, mahler2017dex} for depth map simulation are also motivated by this observation; \\
3) Active stereovision depth sensor has wide adoption. Compared with other kinds of depth sensors, active stereovision depth sensors have relatively high accuracy, high spatial resolution, low cost, and robustness to lighting conditions~\cite{CHEN2022106763}, and therefore are widely used in academic and industry scenarios~\cite{giancola2018survey};\\
4) Active stereovision depth sensor simulation requires the estimation of fewer parameters. Active stereovision depth sensors usually work at a specific wavelength, mostly infrared (IR). In an indoor environment, the energy within the narrow IR band is quite limited, and IR cameras will filter out most of the light in other spectra. Therefore, when modeling the environment, passive environment light is negligible, allowing us to ignore the complex environment illumination, which is the main challenge if we otherwise simulate passive stereovision sensors or RGB sensors;\\ 
5) Real-time realistic ray tracing techniques have matured just recently. Active stereovision sensors use IR lights to probe the environment and form stereo image pairs. State-of-the-art ray tracing pipelines allow us to simulate the transport of IR lights with high fidelity in real-time, which is only available recently with hardware acceleration and learning-based denoising. 

However, depth sensing by active stereovision faces unique challenges relevant to the materials of objects, which needs special attention when building a simulation pipeline. As shown in Fig.~\ref{fig:teaser}, the measurement results on objects with optically challenging materials (\textit{e.g.}, transparent) are inferior, due to the complex lighting effects of the projected pattern on these surfaces. To generate simulation data with a relatively low domain gap, \textit{the simulation process must also create material-dependent error patterns similar to real sensors}.

We explore how to build a light-weight yet effective pipeline to model environments and conduct active stereovision depth sensor simulation that is able to simulate visual observations with a low domain gap. As shown in Fig.~\ref{fig:paper_structure}, the pipeline includes the acquisition of object material parameters, the simulation of the IR pattern projector, the transport of IR lights by ray tracing, the simulation of sensor noise, and lastly, stereo matching, all grounded on the {physics of the process}. 
{We highlight two innovations in the pipeline.} The first is how we acquire object material parameters.
We propose a multispectral loss function to acquire object material parameters, which include visual appearance loss and neural network based perceptual loss to help eliminate the mismatch in brightness and color in both visible spectrum and infrared spectrum. Our second highlighted innovation is in adding textured light support. Existing ray tracing rendering systems usually do NOT support the rendering of textured lights. We include textured light support to simulate the IR pattern projector. In order to achieve real-time simulation of the depth sensor, we further integrate learning-based denoisers into the renderer such that we can generate high-fidelity IR images with hundreds of FPS using a small number of samples per pixel (SPP). {Another time-consuming part is stereo matching. We build a GPU-accelerated stereo matching module consists of common algorithms in real-world depth sensors. The accelerated module can achieve 200+ FPS under common settings while usual CPU implementations can hardly achieve 1 FPS.}

In order to validate the effectiveness of our depth simulator for various down-stream applications in vision and robotics communities, we conducted experiments including generating training data for object detection, 6D pose estimation and robot grasping, and estimating the ranking of pose estimation algorithms.
The proposed depth simulator is able to generate large-scale training data for different perception and policy learning algorithms. It enables the trained algorithms to transfer to the real world without any fine-tuning, even for optically challenging objects whose depth measurements are noisy and incomplete. It can also decrease the cost of  comparing different algorithms' performances.

In summary, the contributions of this work are as follows: 
\begin{itemize}
    \item We present a physics-grounded active stereovision depth sensor simulator that has been validated effective for various applications by real-world experiments. The simulator {has been} integrated with the SAPIEN robot physics simulation platform~\cite{xiang2020sapien} and open-sourced for community usage; 
    \item We build a dataset with precisely aligned simulated data and real data, which can be used to evaluate the transferability of object detection and 6D pose estimation algorithms;
    \item We propose a material acquisition method based on the multispectral matching loss and the pixel-wise alignment between the simulation and the real data. 
\end{itemize}

The paper is structured as follows. In Section~\ref{sec:related_work}, we discuss related works. Section~\ref{sec:preliminaries} introduces the mechanism of active stereovision depth sensor and the concept of Physically Based Rendering. Section~\ref{sec:method-overview} gives an overview of our depth sensor simulation method. Section~\ref{sec:material_acquisition} describes the material acquisition method in detail. Section~\ref{sec:sensor_simulation} presents the active stereovision depth sensor simulation pipeline and key technical contributions.  Section~\ref{sec:system_setup} describes the experimental setup in simulation and the real world. Section~\ref{sec:applications} shows the quantitative and qualitative results of various algorithms on domain transferability for various applications. Section~\ref{sec:ablation_study} validates our design choices through ablation studies. 
Section~\ref{sec:conclusion} concludes the work.

\section{Related work}
\label{sec:related_work}

\subsection{Sim-to-Real}
In computer vision and robotics, using synthetic data for training and transferring model to the real world has been a common approach. A first idea is to use rendered CAD models to train algorithms for solving vision tasks such as view point estimation~\cite{su2015render} and object detection~\cite{sun2014virtual}. Sim-to-real approaches can be divided into two main categories: domain randomization and domain adaptation. Domain randomization applies random augmentation on training data to enforce the learning-based methods to extract task-relevant features and be more robust to data corruption~\cite{tobin2017domain, tremblay2018training, andrychowicz2020learning}. However, the distribution of data augmentation has significant influence on the model's transferability~\cite{vuong2019pick},  and it is difficult to make the augmentation distribution cover that of real data while not being too wide for learning methods. Domain adaptation aims to align data between simulation with real domains through manually-designed rules or learned mapping~\cite{bousmalis2017unsupervised, shrivastava2017learning, bousmalis2018using}. Compared to domain adaptation for RGB images, domain adaptation for depth images is far less studied~\cite{patricia2017deep, martinez2018investigating}. 
{The goal of our work is similar to domain adaptation, which is to close the sim-real domain gap by making the simulated data more realistic. However, we choose to achieve this goal through a physics-grounded simulation method, rather than learning-based domain adaptation.}

\subsection{Depth Sensor Simulation}
{For rasterization and ray-tracing renderers, a \textit{clean} depth map is generated as a by-product. The simplest way to simulate the depth sensor is to just use this depth. But it can not reflect the error characteristics of sensor depth in the real world. Thus, using the clean depth in the training data may harm the sim-to-real performance for learning-based methods.}

{\citet{keller2009real}} applied a physical model that simulates the Time-of-Flight (ToF) phase image generation for depth map synthesis. {\citet{meister2013simulation}} proposed to simulate the multi-path interference of ToF sensors via ray tracing. These two methods are designed for ToF sensors and cannot be used for active stereovision sensors. {\citet{landau2015simulating}} simulated the Kinect depth sensor by perturbing the simulated IR image with empirical noise models and performing stereo matching, and the authors validated the effectiveness on flat surfaces and edges. However, the geometry and material of the objects are not taken into account for simulation. {\citet{planche2017depthsynth}} simulated the depth image from CAD models and validated that their method could be used for learning-based pose estimation and classification on real chairs. However, the depth error caused by specularity and transparency cannot be accurately simulated by rasterization, and they did not perform through studies on the most advanced learning algorithms.

{\citet{bousmalis2017unsupervised}} employed generative adversarial networks (GAN) to generate simulated data and decoupled the task-specific content and domain appearance to improve the generalizability of GAN. 
{\citet{sweeney2019supervised}} learned to predict no-depth-return pixels for depth simulation in a supervised fashion; however, their method to generate the required real training data is time-consuming and prone to error. {\citet{gu2020coupled} used hole prediction network and color-guided GAN to degrade the synthetic depth. \citet{shen2022dcl} proposed to preserve the geometric information in the generated depth by enforcing the differential features to be invariant through the generation process.} One common drawback of these statistical learning depth sensor simulation methods is that the physical mechanisms of light transport, such as specular reflection and refraction, are not taken into account, so they are not able to generate material-dependent depth errors.  {\citet{learningsim2real2019} proposed to search for the optimal sequence of augmentation policies (\eg  white noise, salt noise, etc.) through Monte-Carlo Tree Search (MCTS). However, the complex material-dependent depth errors cannot be imitated by combining simple augmentation policies.} 

{Most recently, \citet{planche2021physics} built an end-to-end differentiable depth simulation pipeline based on differentiable ray-tracing. They optimize the rendering and stereo matching parameters by using the loss between the simulated depth and the real depth.  However, their simulation speed is slow due to current differentiable ray-tracers. Moreover, the differentiable rendering technique they used~\cite{li2018differentiable} does not support rendering of transparent materials, which limits the depth simulation fidelity on these materials. Also, in order to make the pipeline differentiable, the authors build a differentiable stereo matching module, but this pipeline greatly compromises customization capability. For example, it cannot incorporate the directional cost calculation procedure which is widely used in commercial active stereovision depth sensors and can significantly improve the matching quality.}

\subsection{{Depth Completion}}
{Besides depth sensor simulation, depth completion is another technique to improve detecting, grasping, and manipulating optically challenging objects. It aims to improve the quality of captured incomplete and noisy depth maps~\cite{xie2022recent}. ClearGrasp\cite{sajjan2020clear} used the RGB image to predict surface normals, occlusion boundaries, and object masks before using global optimization to reconstruct the depth map. However, global optimization is impractical for real-time robot applications due to its high time cost. \citet{zhu2021rgbd} proposed to use a local implicit neural representation based on ray-voxel pairs to complete the depth and then iteratively refined the depth estimation. To accelerate the computation, they developed a customized CUDA kernel, which requires advanced GPU programming skills. \citet{xu2021seeing} proposed to combine depth completion with object point cloud completion to better utilize the geometry priors and achieved improved performance. But it requires object masks for point cloud completion, which may be unavailable in certain scenarios. TransCG~\cite{fang2022transcg} constructed a large-scale real-world dataset for transparent object depth completion and trained a 2D CNN-based depth completion network on the dataset. However, constructing the real-world dataset with ground-truth depth and expanding it to include novel objects required a considerable amount of time. }

{Although depth completion methods are able to complete the depth map, they do not take into account the noise pattern in measurable areas and may introduce additional artifacts, limiting their ability to narrow the domain gap. Moreover, depth completion increases the computation cost when used for downstream tasks. In contrast, depth sensor simulation is able to improve the network's performance on real corrupted depth maps without introducing any additional burden at inference time. }

\subsection{Material Acquisition}
\label{subsec:rel-material}
The goal of material acquisition is to estimate material properties from images with known geometry, and is a long-standing problem in the field of computer graphics. To adequately sample the surface function, classical methods typically require complex controlled camera and lighting setups \textit{e.g.} light stage~\cite{debevec2000acquiring} or structured lights~\cite{gardner2003linear, ghosh2009estimating} . A few recent methods utilize neural network to extract bidirectional reflectance distribution function (BRDF) maps from captured images, and have enabled material acquisition from more casual setups such as flash photographs~\cite{aittala2016reflectance,li2018materials,henzler2021generative}. More recently, \cite{zhang2021nerfactor} proposed employing a neural radiance field~\cite{mildenhall2020nerf} based pipeline to jointly recover a fully-factorized 3D model from images captured under a single illumination. These methods represent materials in the form of spatially-varying BRDF (SVBRDF), which typically includes diffuse albedo, normal, roughness and metallic maps. The majority of these methods do not consider transparent materials. Consequently, they cannot be applied to many common real objects such as bottles and glasses. Our setting differs from these previous work in that we assume a known CAD model with a diffuse base color, which is very common in the field of robotics. With {this} information, we can assume a physically based rendering (PBR) material for each object part and search for material parameters efficiently.

\subsection{Robot Simulation}
Thanks to the recent progress of physics simulation~\cite{todorov2012mujoco, coumans2016pybullet, erez2015simulation}, it has drawn increasing interest to build full-physics robot simulation environments~\cite{urakami2019doorgym, tunyasuvunakool2020dm_control, zhu2020robosuite, james2020rlbench, mu2021maniskill}. Compared to robot simulation with abstract action~\cite{kolve2017ai2, savva2019habitat, xia2020interactive}, full-physics robot simulation supports low-level policy learning that could be transferred to the real world. 
Therefore, we integrate our depth simulator with state-of-the-art robot simulator SAPIEN~\cite{xiang2020sapien}, targeting at vision-based robot policy learning. To the best of our knowledge, our work is the first physics-grounded depth simulator integrated into a robot simulation environment.

\section{Preliminaries}
\label{sec:preliminaries}
\newcommand{\nbf}[0]{\mathbf{n}}
\newcommand{\ibf}[0]{\mathbf{i}}
\newcommand{\obf}[0]{\mathbf{o}}

\subsection{Active Stereovision Depth Sensor}
Existing commercial optical depth sensors can be broadly categorized into four groups: passive stereovision, active stereovision, structured light, and Time-of-Flight (ToF). In this section, we will introduce the mechanism of active stereovision depth sensors.
We refer to \cite{giancola2018survey} for a detailed introduction of the principles of other depth sensing techniques. 

\begin{figure}
    \centering
    \includegraphics[width=0.8\linewidth]{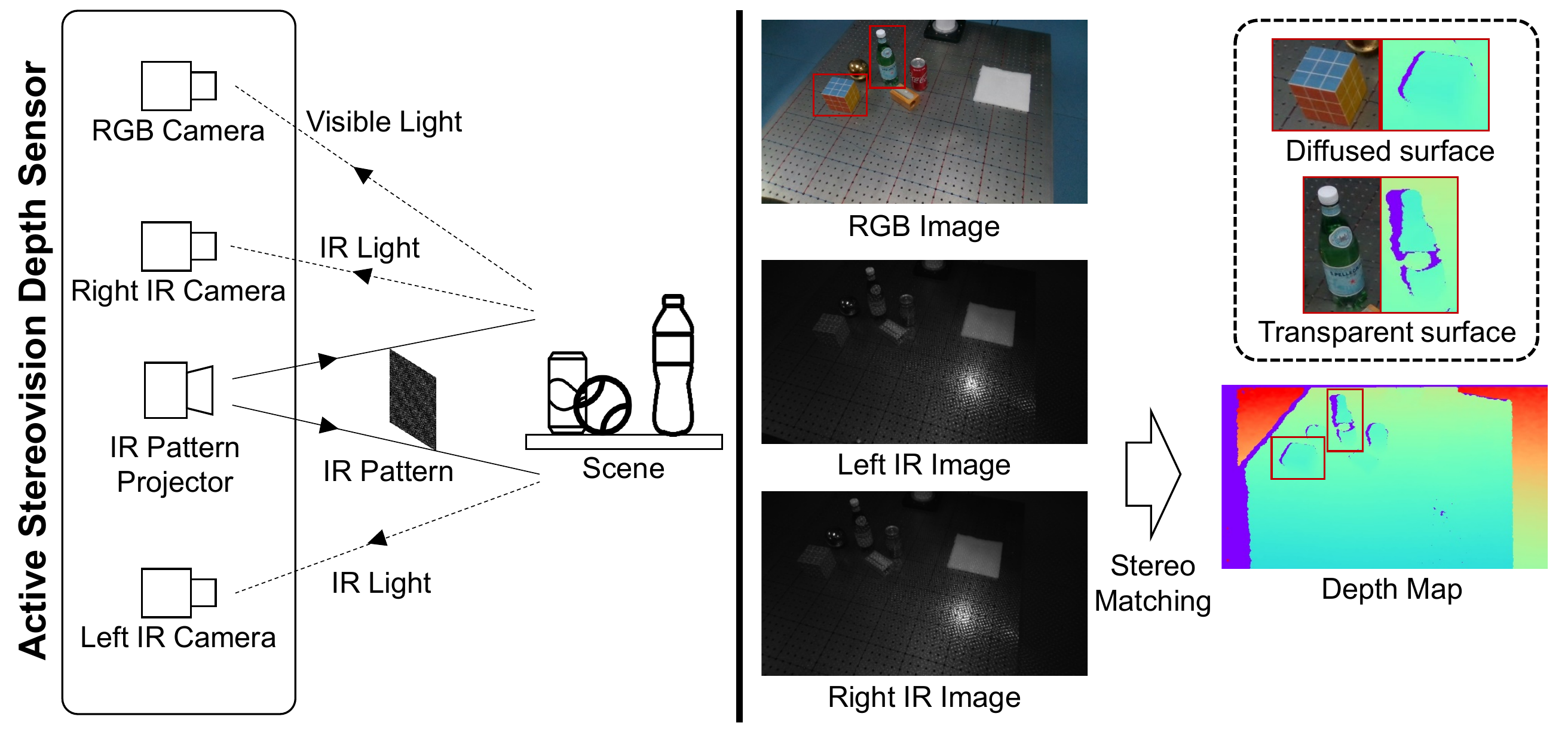}
    \caption{Mechanism of active stereovision depth sensors{.}}
    \label{fig:active_stereovision}
\end{figure}

As shown in Fig.~\ref{fig:active_stereovision}, an active stereovision depth sensor consists of an IR pattern projector, two IR cameras, and an RGB camera. All transformations between the three cameras are known. The projector casts an IR pattern (typically random dots) onto the scene. The two IR cameras capture the scene with the reflected pattern in the IR spectra. The depth map is computed by stereo matching on the two captured IR images. One common post-process is to warp the depth map to the RGB image from the RGB camera to generate aligned RGBD data. 
Compared with passive stereovision, active stereovision can measure texture-less areas. Compared with structured light which uses coded pattern(s), active stereovision is unaffected when multiple devices measure the same area simultaneously. Compared with ToF, active stereovision has higher spatial resolution and is not affected by multi-path interference.

While an active stereovision depth sensor can achieve satisfactory performance on objects with diffused surfaces where the projected pattern is uniformly reflected, the results on objects with challenging materials (\textit{e.g.}, glossy or transparent) are inferior due to the complex lighting effects of the projected pattern on these surfaces. To generate simulation data with a relatively low domain gap, the simulation process must also generate the same material-dependent measurement error patterns as real sensors.

\subsection{{Physically Based Rendering}}
\label{sec:pre_pbr_material}
Physically based rendering (PBR) seeks to render images in a way that models the transport of light in the real world. In nature, a light source emits light rays that travel to surfaces interrupting their progress. Depending on the surface material, the rays could get absorbed, reflected, refracted, or any combination of these. A PBR pipeline seeks to model both the light transport process and the surface material in a physically correct way.

\textbf{Light transport process.} Many rendering techniques have been proposed to model the light transport process, such as ray casting, recursive path tracing, and photon mapping. These methods are covered by an umbrella term \textit{ray tracing}. In general, a ray tracing algorithm calculates the color for each pixel by tracing a path from the camera and accumulates the material-dependent weight along the path to light sources. Compared to the traditional scan-line-based renderer (rasterizer), ray-tracing-based renderers (ray-tracers) are typically slower but can produce much more photo-realistic images with complex indirect lighting effects. They have shown promising sim-to-real transferability on vision tasks, \textit{e.g.}, object recognition ~\cite{zhang2017physically, hodavn2019photorealistic}. Recent years have seen great progress on accelerating ray tracing from both academia and industry. The adoption of deep-learning-based denoising~\cite{bako2017kernel,chaitanya2017interactive} and super-resolution~\cite{dong2015image} has made real-time ray tracing possible. In this work, we implement a new ray-tracing-based renderer that integrates many of these cutting-edge acceleration approaches and customizable features to support our depth sensor simulation pipeline.

\begin{figure}
    \centering
    \includegraphics[width=0.3\textwidth]{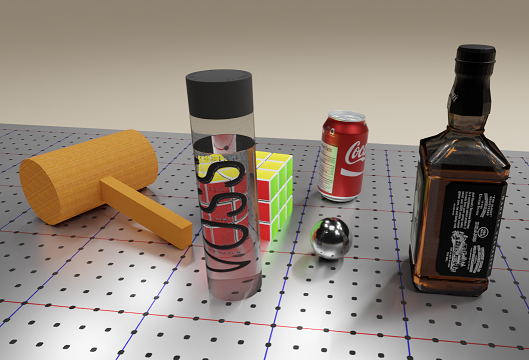}
    \caption{Ray-tracing rendering result of a table-top scene. Our renderer features broad support for physically-based rendering materials and can produce realistic indirect lighting effects (\textit{e.g.} soft shadow, reflection, and refraction).}
    \label{fig:render-demo}
\end{figure}

\textbf{Surface material modeling.} {Another crucial factor in producing photo-realistic images, as aforementioned, is the physically correct modeling of surface materials and light transportation. In the real world, the BSDF (Bidirectional Scattering Distribution Function) describes how light scatters from a surface. It is defined as the ratio of scattered radiance in a direction $\obf$ (outcoming light direction) caused per unit irradiance incident from direction $\ibf$ (incoming light direction). We denote it as $f(\ibf,\obf,\nbf)$ where $\nbf$ denotes the local surface normal. The BSDF is often called BRDF or BTDF when restricted to only reflection or transmission.}
The BSDF of a real-world surface is very complex. However, researchers have proposed many parameterized models to approximate the function. Among all these efforts, the Disney\texttrademark~PBR material model~\cite{burley2012physically} (referred to as PBR material) is most widely recognized. Based on a large-scale investigation of various real materials, the model parameterizes the BSDF into a set of parameters including base color, metallic, specular, roughness, anisotropic, sheen, clearcoat, transmission, emission, etc. This model is widely adopted by industry-level softwares such as Blender\texttrademark, Renderman\texttrademark~, and Unreal Engine\texttrademark. {A detailed demonstration of the effect of these parameters can be seen in Blender Principled BSDF Introduction\footnote{https://docs.blender.org/manual/en/latest/render/shader\_nodes/shader/\\principled.html}. In our rendering solution, we have adopted a version of the PBR material model that includes base color, metallic, specular, roughness, index of refraction (IOR), transmission, and emission as its parameters. These parameters suffice to model most common materials in the real world (see Fig.~\ref{fig:render-demo} for examples). Many recent works have employed a PBR material~\cite{shen2021igibson,xia2018gibson,kolve2017ai2}. But they all neglect some of the parameters (metallic, transmission, or emission). Combined with their rasterization-based renderers, this could lead to unrealistic results for some common household objects such as lamps, mirrors, and glasses under complex lighting conditions.}

\section{Method overview}
\label{sec:method-overview}

In this section, we give a more detailed description of our objective and our method to achieve it.

The objective of our work is to close the optical sensing domain gap between the depth sensor simulator and the real depth sensor. The depth sensor simulator is defined as:
\begin{equation}
    I_{\textrm{depth}} = D_{\psi}(\mathcal{S}_\textrm{objects}, \mathcal{S}_\textrm{lighting})
\end{equation}
where $ I_{\textrm{depth}}$ is the output depth map, $\psi$ is the depth sensor simulator parameters, $\mathcal{S}_\textrm{objects}, \mathcal{S}_\textrm{lighting}$ are the object status and the lighting status in the robot simulation environment. $\mathcal{S}_\textrm{objects}$ includes the geometries, PBR materials (base color, metallic, specular, roughness, transmission, IOR, emission), and poses of objects. $\mathcal{S}_\textrm{lighting}$ includes the positions and intensities of light sources.
In this paper, we assume the geometries and base colors of objects are available from textured CAD models\footnote{For example, if we use a multi-view stereo 3D scanner to reconstruct the surface of 3D objects, we will acquire such textured CAD models.}. The positions of light sources are approximated in the real world. This is a common and useful setup in robotic research. In many scenarios, such as logistics, a collection of CAD models can be build beforehand for all the objects of interest.

{The optical sensing domain gap can be evaluated as follows: a group of perception models are evaluated on one simulation dataset and one real dataset, which are accurately aligned. The optical sensing domain gap is considered smaller if the evaluation results on the simulation dataset are more correlated with those on the real dataset. Moreover, the effectiveness of simulation as the training domain is reflected by the algorithm performance in the real world.}

In order to close the optical sensing domain gap, we will 1) acquire the unknown PBR materials of objects, including metallic, specular, roughness, and transmission, along with the intensities of light sources; and 2) construct $D_{\psi}()$ by simulating an active stereovision depth sensor in a physics-grounded fashion.

\section{Material acquisition}
\label{sec:material_acquisition}
Synthesizing data from textured CAD models is a common practice in previous robotic works~\cite{calli2017yale, qin2020s4g,liu2021ocrtoc}. Almost all of these works assume a simple material model (\textit{e.g.}, Lambertian) and do not set the material parameters carefully. This is generally acceptable when using a rasterization-based renderer, which inherently cannot generate indirect lighting effects such as specular reflection, refraction, and shadows, without additional tricks. Because our goal is to count in all real-world factors including these lighting effects, a ray-tracing renderer along with a good PBR material model are needed.

In order to acquire the object material parameters, we first develop a labor-saving and markerless pipeline to achieve pixel-wise alignment between the simulation and real scenes. Furthermore, with the scene setting aligned, we can achieve material acquisition by optimizing PBR material parameters for each object part.

\subsection{Pixel-wise Alignment between Simulation and Real World}
\label{sec:pixel_wise_alignment}

To retrieve the materials of objects, it is necessary to accurately align the scene configuration, including camera parameters, object geometries, and poses, between the simulation and the real world so that the rendered images are pixel-wise aligned with the real-world captures. Previous works achieved this goal by estimating the object poses in the initial camera frame, using markers (\textit{e.g.}, QR code cardboards) to track the pose of camera frames, and computing the object pose for each frame~\cite{rennie2016dataset, xiang2017posecnn}, but it is undesirable to leave markers in captured images. We eliminate the need for markers by first generating the rendered images and then aligning the real scene layout with the simulation instead of aligning the simulation with the real world.

\begin{figure}[t]
    \centering
    \includegraphics[width=0.8\linewidth]{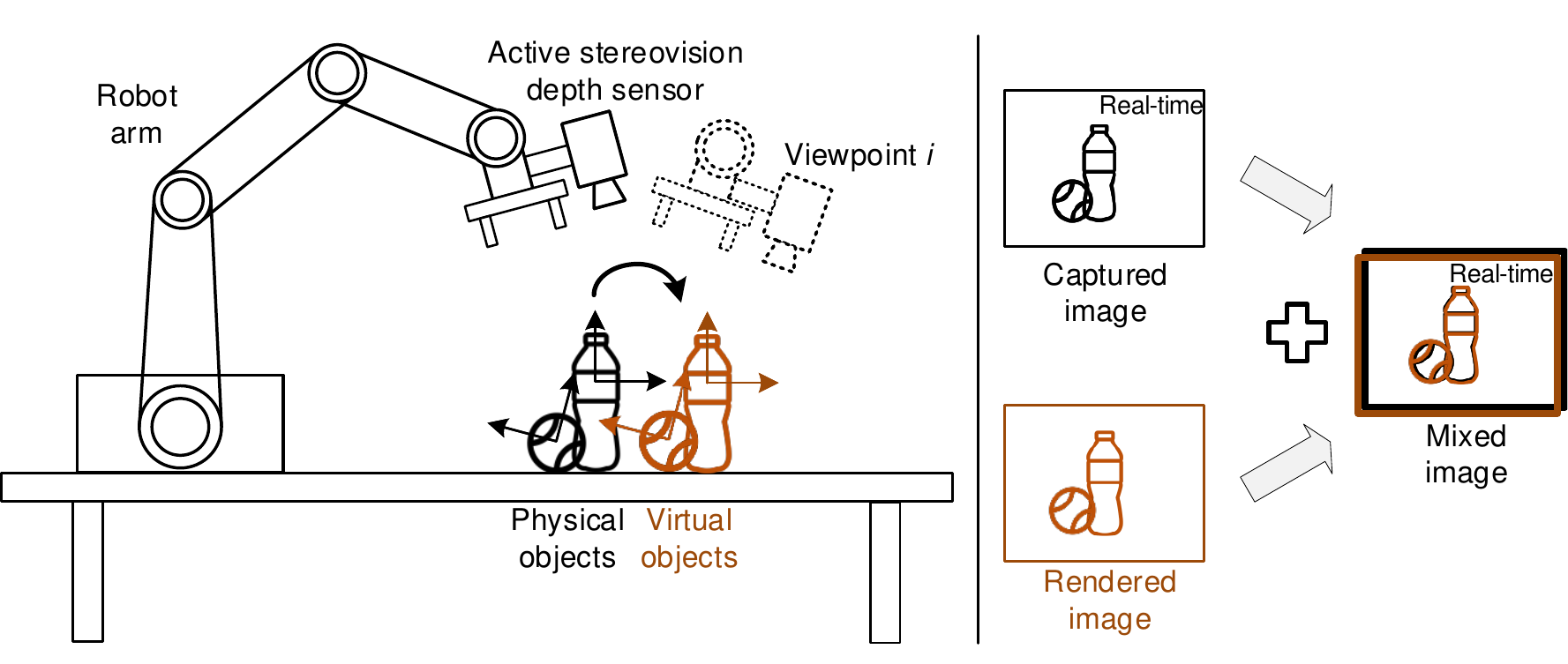}
    \caption{Demonstration of setup for material acquisition, and pixel-wise alignment between the simulation and the real world.}
    \label{fig:alignment_setup}
\end{figure}

Fig.~\ref{fig:alignment_setup} shows the setup for material acquisition. The active stereovision depth sensor is fixed at the robot end effector. Objects are placed on the table, and the depth sensor captures them from different angles. 
To align the simulation environment with the real world, we need to compute the relative transformations for depth sensors, the robot, and the table in the following steps: 1) the camera intrinsic matrices of the depth sensor are obtained from the firmware; 2) the relative transformations between cameras and the table of all the viewpoints are computed using OpenCV's \texttt{solvePnP}~\cite{bradski2000opencv,terzakis2020consistently}; and 3) the relative transformation between the depth sensor and the robot end effector is computed using hand-eye calibration from OpenCV~\cite{daniilidis1999hand}. Once all the relative transformations are obtained, we use the aligned poses of the depth sensors, the robot, and the table to construct the simulation environment and maintain them across all scenes.

After building the simulation environment, we proceed to create aligned object layouts between the simulation and the real world. We first use the physics simulation to generate a physically plausible scene layout of objects in simulation following the pipeline in \cite{qin2020s4g}. Secondly, we render images of the scene. With well-aligned transformations between the simulation and the real world, the pose of the table is precisely aligned in rendered images and captured images. As shown in the right part of Fig.~\ref{fig:alignment_setup}, we generate the mixed image in real time by overlaying the rendered image on the captured image as the feedback, and adjust the object poses according to this real-time feedback without the aid of markers. Because the layout is physically plausible, we can accurately align the object poses in the real world with those in the simulation. {We measure the displacement accuracy using a motion capture system, and the result demonstrates that we can achieve an error of less than 2mm, $1.5\degree$, which is sufficient for the evaluation of pose estimation.}

\subsection{Material Acquisition}

With the scene layout already aligned between the simulation and the real world, we can now retrieve the material parameters of objects in the scene. Note that we need approximated positions of main light sources in this step ($\sim10\text{cm}$ accuracy in our experiments), and the light intensities will be optimized automatically along with the material parameters. We first pre-process the real captured images using automatic exposure and white-balancing to align the image brightness and color, and then we search through light intensities and several important PBR material parameters, including roughness, metallic, specular, and transmission, for each object part to find the best match. 

\textbf{Multispectral matching Loss.} For an aligned pair of multispectral images $(I_\text{RGB}^\text{sim}, I_\text{IR}^\text{sim})$ and $(I_\text{RGB}^\text{real}, I_\text{IR}^\text{real})$, the matching loss $L$ is defined as:

\begin{align}
    L &= L_{\text{RGB}} + \lambda L_{\text{IR}},\\
    L_{\text{RGB}} &= L_2(I_{\text{RGB}}^{\text{sim}}, I_{\text{RGB}}^{\text{real}}) + L_{percept}(I_{\text{RGB}}^{\text{sim}}, I_{\text{RGB}}^{\text{real}}), \\
    L_{\text{IR}} &= L_2(I_{\text{IR}}^{\text{sim}}, I_{\text{IR}}^{\text{real}}) + L_{percept}(I_{\text{IR}}^{\text{sim}}, I_{\text{IR}}^{\text{real}})
\end{align}
where $L_{percept}$ is the perceptual loss used to match the overall appearance of the objects. The perceptual loss is defined as the $L_2$ difference between AlexNet features of the input {rendered 2D images}~\cite{johnson2016perceptual}. We empirically find that by adding the perceptual loss, we can achieve better material acquisition results with no need for strict color, exposure and lighting condition alignment, while plain $L_2$ could give sub-optimal results due to bias on brightness and color. 

\textbf{Adaptive grid search.} Thanks to the real-time performance of our ray-tracing pipeline, we can perform a grid search on the PBR material parameters for each object part to find the optimal match. The parameter set $\mathcal{P}$ we search for includes roughness, metallic, specular, and transmission. We first perform a coarse-level grid search with 10 grid samples for each parameter. After obtaining the best set $\mathcal{P}^{\text{coarse}}$, we perform another fine-level grid search with 10 grid samples in the neighborhood of $\mathcal{P}^{\text{coarse}}$ to acquire the final material parameter approximation $\mathcal{P}^{\text{fine}}$. {These identified material parameters are then applied to 3D models and used to render images.}

\section{Active stereovision depth sensor simulation}
\label{sec:sensor_simulation}
\newcommand{\mbf}[0]{\mathbf{m}}
\newcommand{\hbf}[0]{\mathbf{h}}

In this section, we introduce individual components of our end-to-end solution for active stereovision depth sensor simulation.

This proposed pipeline follows the same underlying mechanism of real active stereovision depth sensors. Specifically, we simulate the full processing flow from IR pattern projection and rendering to IR noise simulation and depth reconstruction. We build our pipeline based on SAPIEN~\cite{xiang2020sapien} as it already has state-of-the-art physical simulation and user-friendly robotic interfaces integrated so that we can focus on the sensor simulation.

\subsection{Real-Time Physically Based Rendering}
\label{subsec:pbr}
Originally, SAPIEN features a real-time rasterization renderer, which meets the demands of many simple robotic and vision tasks that do not require a high degree of realism. However, complex optical effects such as indirect lights, soft shadows, cannot be easily rendered by such a rasterization-based pipeline, leading to unrealistic results.

Building upon the SAPIEN platform, we add a new ray-tracing renderer named \textit{Kuafu} in this paper. Kuafu is implemented with the Vulkan ray-tracing API, which provides real-time performance as well as cross-vendor and cross-platform compatibility. Kuafu renderer can generate physically correct realistic rendering results (Fig.~\ref{fig:render-demo}) and can render in real time for common robotic scenes under reasonable settings. In addition, Kuafu supports many advanced render functionalities, including geometry instancing, environment map, and customizable render target. To support the realistic sensor simulation pipeline, {we highlight the following features:}

\textbf{PBR material support.} Kuafu adopts a subset of the widely-used Disney\texttrademark~PBR material model~\cite{burley2012physically}. For the shader implementation, we follow these existing frameworks to use the Lambertian model or (optional) the Oren-Nayar model~\cite{wolff1998improved} for diffused component modeling and GGX microfacet model~\cite{walter2007microfacet} for specular component modeling. Our material model includes base color, metallic, specular, roughness, index of refraction (IOR), transmission, and emission. All parameters can be either from a single scalar or from an assigned image texture, giving it great flexibility to model all common object materials. It is worth noting that we feature full support for transmissive and emissive materials, which is uncommon in previous simulation platforms, however supporting {these} materials is crucial for accurate sensor simulation, as will be shown in later sections.

{The final BSDF model we use is a mix between the diffuse term and the specular term:
\begin{align}
    f(\ibf,\obf,\mbf) &= w_d f_d(\ibf,\obf,\mbf) + (1 - w_d) f_s(\ibf,\obf,\mbf), \\
    w_d &= (1 - \mu) * (1 - \alpha),
\end{align}
where $\mu$ is the material metallic, $\alpha$ is the object transmission, $f_d$ is the diffuse term (Lambertian or Oren-Nayar) and $f_s$ is the specular term (GGX microfacet model). $\ibf,\obf,\mbf$ denote unit incoming light direction, outcoming light direction and local surface normal, respectively. }

\textbf{Textured light support.} Kuafu supports a wide variety of light types, including directional, point, spot, and area lights. Most lights in Kuafu feature texture map support. This is implemented in shaders by first determining which pixel of the light texture the ray passes through when tracing shadow rays, and then attenuating the light intensity according to the light texture. This feature further improves flexibility and extends to a broad range of usages. For instance, combined with the variable texture support, one can use lights with programmed textures to simulate the shape from the shading pipeline~\cite{zhang1999shape} by changing the light pattern across the frames. Also, the pattern projector in active stereovision sensors can be simulated by a textured spot light in Kuafu, and thus this functionality is crucial for our proposed simulation pipeline.

\textbf{Denoising support.} Images from a vanilla ray tracing pipeline typically contain significant noise due to the underlying Monte Carlo sampling process. The noise is especially severe when the scene is under-illuminated or rendered with a small number of samples per pixel (SPP), both of which can be encountered when rendering IR images. To reduce these undesired noises while using smaller SPPs, we integrate two deep-learning-based denoisers into the renderer -- NVIDIA\texttrademark~OptiX Denoiser~\cite{parker2010optix} and Intel\texttrademark~Open Image Denoise. By explicitly sharing GPU memory between Kuafu and the OptiX CUDA buffer, we eliminate user-space memory copying and thus achieve high-quality denoising with a minimal performance impact. As showed by \citet{chaitanya2017interactive}, deep-learning-based denoisers can achieve visual appearance very similar to physically-based rendering. Therefore, we feel it reasonable that our work can still be summarized as physically grounded. 

\subsection{Infrared Image Rendering}
With the support of textured spot light, it is straightforward to render IR images in Kuafu. In our implementation, when rendering IR images, we assume that all existing lights in the environment will also cast rays in the IR spectrum, with reduced intensity $a*l$, where $l$ is the light intensity in the visible spectrum and $a$ is an attenuating factor set to $0.05$ in all our experiments. We also assume a weak ambient light value simulating radiance from the environment (\textit{e.g.}, sunlight). We render images with this special light setting, and then take the R channel of the denoised images as the final IR rendering results. The rendering speed of an IR image is shown in Table~\ref{tab:performance}. 

{There are indeed commercial physically-based solutions like Blender and 3dsMax which feature a similar full ray-tracing pipeline. However, these solutions are often heavy and could bring in a large amount of run-time overhead. By implementing a similar full-featured pipeline ourselves, we can: 1) achieve performance gains by optimizing out the unnecessary steps; 2) implement certain techniques for better efficiency and performance (e.g. sharing GPU memory between Kuafu and the stereo matcher, and customize on-device post-processing); 3) build a unified pipeline into robotics simulators. Users can use the Kuafu renderer and depth simulator out-of-the-box in our SAPIEN robotics simulator with only a few lines of code.}

\begin{table}[h]
\caption{{Rendering frame rate (FPS) with different numbers of sample per pixel (SPP). The experiments are conducted on a single RTX 4090 graphics card. The rendering resolution is $960\times540$.}\label{tab:performance}}
\centering
\begin{tabular}{c|c|c} 
  \toprule
  Sample Per Pixel  &  w/o Denoiser & w/ Denoiser \\
  \hline
  2    &352.80  &140.63  \\ 
  8    &146.51  &100.47  \\ 
  32   &57.36   &43.70  \\ 
  128  &27.93   &24.79  \\ 
  \bottomrule
\end{tabular}

\end{table}

\subsection{Sensor Noise Simulation}
\label{sec:method_sensor_noise}
After the IR projection, rendering, and denoising steps, we can acquire clean IR images with complex indirect lighting effects. However, real-world cameras and projectors are imperfect. Due to hardware or manufacturing limitations, real-world cameras and projectors will introduce different kinds of noise into IR images.

We generally follow the model used in~\cite{landau2015simulating} to model the IR sensing noise. After obtaining the rendered IR image $I_\text{clean}$ (in light intensity), we corrupt the image with random noise to simulate the noise encountered in real optical systems. Our noise model contains two parts: a multiplicative term $\gamma$ modeling the laser speckle and an additive term $n$ modeling camera thermal noise: 
\begin{equation}
I_\text{noisy} = \gamma * I_\text{clean} + n,
\end{equation}
where $*$ and $+$ denote element-wise operations. According to \cite{goodman2007speckle}, the total power received by a pixel can be modeled as the sum of exponentially distributed power (Rayleigh voltage) random variables. As a result, we can fit a gamma distribution to sample this noise:
\begin{equation}
f_\gamma(x; k, \theta) = \frac{1}{\Gamma(k)\theta^k}x^{k-1}e^{-\frac{x}{\theta}},
\end{equation}
where $k$ and $\theta$ are two noise parameters. The camera thermal noise can be modeled by a {Gaussian distribution $\mathcal{N}(\mu,\,\sigma^{2})$~\cite{perepelitsa2006johnson}.}

{For a specific sensor that we choose to simulate, we estimate the parameters of its noise model by capturing multiple sets of IR images, where each set contains frames of the same static scene. In this way, we can calculate the noise-free approximation of each scene by averaging the frames, and then estimate the noise model parameters with a maximum a posteriori probability (MAP) estimate.}

\subsection{Depth Generation by Stereo Matching}
{After noise simulation, a series of computationally expensive stereo matching steps need to be performed to generate the final depth map. Usual CPU implementations have a computation time that is far from real-time. To accelerate our pipeline, we build a high-speed GPU-based depth computing module named \textit{SimSense}. SimSense is an encapsulated module that integrates all necessary algorithms to compute depth from a pair of stereo images. It is highly configurable and has been fully accelerated with CUDA.}

{With a pair of stereo images as input, SimSense first performs a stereo rectification to project the images onto a common image plane. It then performs a center-symmetric census transform (CSCT)~\cite{zabih1994non} on the input as suggested in the RealSense overview~\cite{8014901}. After CSCT, a four-path semi-global matching (SGM)~\cite{hirschmuller2007stereo} is performed to search for the best disparity candidates, with hamming distance as the cost function. A uniqueness test is done to filter out disparities that are not better than the second best match by a threshold. Sub-pixel-level disparity is generated by performing quadratic curve fitting. Several common disparity post-processing steps including left-right consistency check and median filtering can be done to refine the disparity map. The refined disparity map is then converted into a depth map, with an optional depth registration that aligns the depth map to the RGB camera frame.}

{We follow the parallel scheme proposed by Hernandez-Juarez et al.~\cite{HERNANDEZJUAREZ2016143} to accelerate CSCT and SGM. CSCT and matching cost computation of SGM make use of shared memory to optimize for data reuse. Cost aggregation of SGM is further accelerated with warp-based optimization. Rectification, uniqueness test, sub-pixel interpolation, left-right consistency check, median filter, and depth registration are integrated within the parallel scheme with little overhead. Besides the original SGM, SimSense also supports semi-global block matching (SGBM). While SGM computes the matching cost of two pixels solely by their hamming distance, SGBM computes the cost by the hamming distance between the local regions of the two pixels. Such implementation is common in library like OpenCV~\cite{bradski2000opencv}. Table~\ref{tab:simsense} shows the computation time performance of SimSense with a comparison to the CPU implementation of SGBM with OpenCV.}

{Key parameters are designed to be adjustable throughout the pipeline. Thanks to the configurability of the module, the entire system can easily adapt to new sensors.}

\begin{table}[h]
\caption{{Depth computing frame rate (FPS) under with different max disparity values. Our SimSense algorithms run on a single RTX 4090 graphics card. The OpenCV implementation is run on a machine with i5-10400 CPU @ 2.90GHz, and 16G memory. The input image resolution is $960\times540$ and the generated depth map resolution is $1920\times1080$.}\label{tab:simsense}}
\centering
\begin{tabular}{c|c|c|c} 
  \toprule
  Max Disparity & SGM (Ours) & SGBM (Ours) & SGBM (OpenCV) \\
  \hline
  64    &414.51   &342.24   & 0.73 \\ 
  96    &326.79   &268.98   & 0.71 \\ 
  128   &281.26   &232.49   & 0.68 \\ 
  256   &147.13   &128.56   & 0.60 \\
  \bottomrule
\end{tabular}
\end{table}

\section{System setup}
\label{sec:system_setup}
\subsection{Experimental Setup}

\begin{figure}
    \centering
    \includegraphics[width=0.8\linewidth]{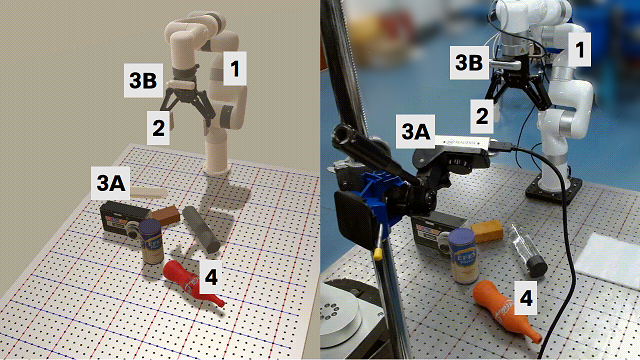}
    \caption{Experimental setup: (1) 7-DoF robot arm, (2) parallel gripper, (3A) Intel RealSense D415 mounted on the table { (}base RealSense), (3B) Intel RealSense D415 mounted on the robot end effector { (}hand RealSense), (4) Target Objects{.}}
    \label{fig:experiment_setup}
\end{figure}

\begin{figure}[tb]
    \centering
    \includegraphics[width=0.9\linewidth]{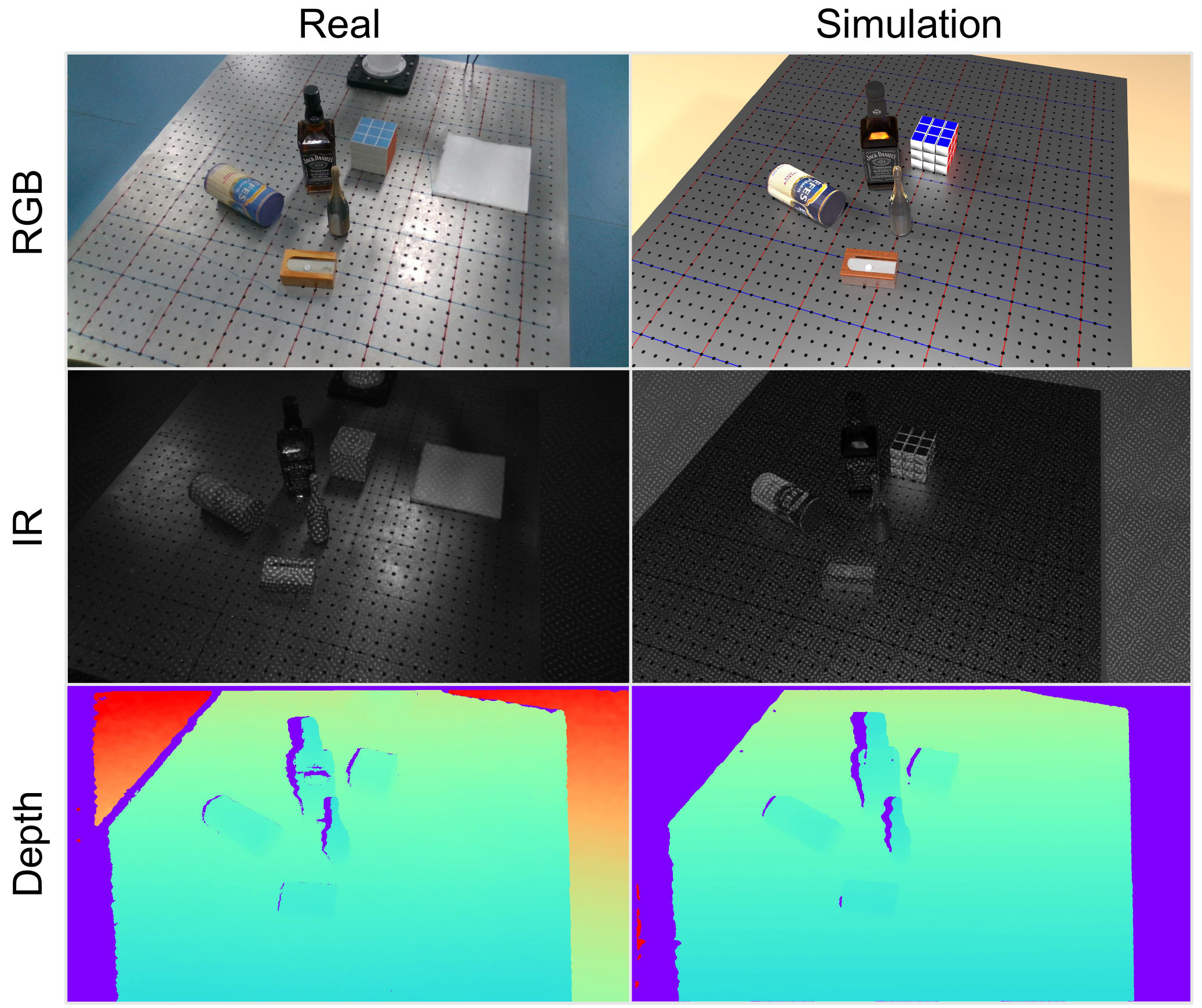}
    \caption{Pixel-wise alignment between simulation and the real world.}
    \label{fig:alignment}
\end{figure}

Fig.~\ref{fig:experiment_setup} shows the experimental setup in simulation and the real world. It consists of a 7 DoF robot arm (UFACTORY xArm 7), a parallel gripper (Robotiq 2F-140), 2 IR active stereovision depth sensors (Intel RealSense D415), and objects. One of the depth sensors is fixed on the table (referred to as \textit{base RealSense}) and the other is mounted to the robot end effector (referred to as \textit{hand RealSense}). We further build the simulation environment in SAPIEN~\cite{xiang2020sapien} and align the real world and simulation as described in Section~\ref{sec:pixel_wise_alignment}. Fig.~\ref{fig:alignment} shows the alignment result.

\subsection{Objects and Material Acquisition}
\label{sec:objects}

\begin{figure}
    \centering
    \includegraphics[width=\linewidth]{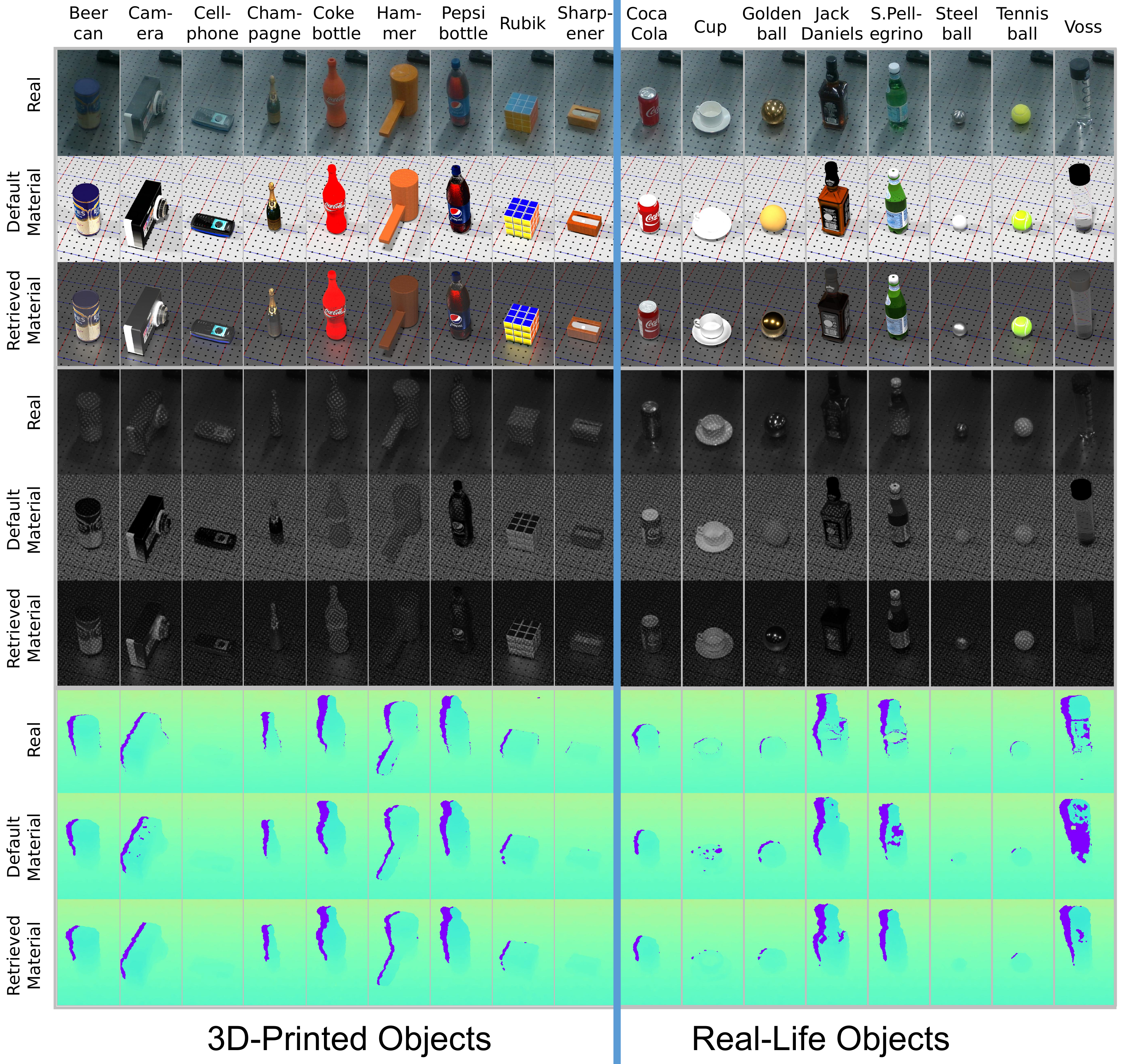}
    \caption{Comparison of captured and rendered RGB images, IR images and depth maps of 3D-printed objects and real-lift objects{.}}
    \label{fig:objects_print}
\end{figure}

As shown in Fig.~\ref{fig:objects_print}, there are two kinds of objects used in our experiments: 9 3D-printed objects and 8 real-life objects. 3D-printed objects are printed using color plaster powder from CAD models, therefore the geometry and texture consistency between physical objects and CAD models is guaranteed and researchers can print these objects by themselves. {Note that ``texture'' here only refers to surface reflectance, also known as albedo or diffuse color of an object.} Because the materials of 3D-printed objects are all Lambertian diffused, we further choose 8 real-life objects of different optical materials (specular, translucent, transparent) which are challenging for active stereovision depth sensors. {We approximate the material parameters of all the objects by conducting material acquisition as described in Section~\ref{sec:material_acquisition}. For each object, it takes less than 1 second to render the synthetic images for object pose alignment. And it takes about 2 minutes to control the robot arm to move the camera around the object and capture images in the real world. Finally, it takes less than 1 hour to search for the best material parameters for an object.}

\subsection{Sensor Parameters}

For our Intel RealSense D415, to estimate the parameters of our noise model as described in Section~\ref{sec:method_sensor_noise}, we capture 33 sets of IR images, where each set contains 100 frames of a static scene. The estimated parameters are $\mu=-0.231, \sigma=0.83, k=3.98, \theta=0.254$. To retrieve the IR projector pattern of D415, we capture a perpendicular wall multiple times to filter out noise and then use simple thresholding to extract the pattern.

\section{Domain Transferability Evaluation on Diverse Applications and Algorithms}
\label{sec:applications}
We define four applications of the proposed depth sensor simulator to evaluate the domain transferability.

\textbf{1. Generate training data for object detection}
Object detection is one of the most widely deployed perception algorithms.
{In Section~\ref{sec:application_object_detection}, we generate a training dataset using the proposed depth sensor simulator. On the simulated training dataset, learning-based object detection algorithms are trained before being evaluated on the real-world dataset. }

\textbf{2. Generate training data for 6D pose estimation} 
Pose estimation is critical for downstream robotic applications (pick-and-place, manipulation). Compared with object detection, it is more difficult to manually label accurate annotations of 6D poses.
{In Section~\ref{sec: application_pose_estimation}, we choose three learning-based 6D pose estimation algorithms and run them through the same pipeline as object detection. }

\textbf{3. Generate training data for robot grasping}
In contrast to object detection and 6D pose estimation, where annotations can be generated simply from the state of the static virtual scene, the annotations of robot grasping require physical interaction with the environment. To do this, we integrate the proposed depth simulator with the full-physics robot simulation environment SAPIEN~\cite{xiang2020sapien}. In Section~\ref{sec:application_robot_grasping}, we train a continuous robot control policy in the simulation environment, and directly test the policy in the real world. 

\textbf{4. Estimate the ranking of pose estimation algorithms in real world}
{In addition to generating training data, the proposed depth sensor simulator can also be used to rank algorithms. It facilitates researchers to compare different algorithms in simulation without conducting experiments in the real world.
In Section~\ref{sec:application_algorithm_benchmarking}, we generate a simulated dataset using our proposed depth sensor simulator, and align it with a real-world dataset. Different pose estimation approaches are assessed on the datasets simulated by our simulator and other approaches. We compare the rankings of  algorithms on simulated datasets with rankings on the real-world dataset.}

\subsection{Generate Training Data for Object Detection}
\label{sec:application_object_detection}

\textbf{{Training dataset.}}
{
For both learning-based object detection and pose estimation, we generate a large-scale synthetic training dataset. As shown in Fig.~\ref{fig:experiment_setup} and Fig.~\ref{fig:alignment}, the scene in the dataset is constructed by randomly choosing 5 objects and placing them on the table, and depth maps are generated by simulating the depth sensor from different viewpoints. In order to avoid generating unreasonable object layouts, we use physics simulation to generate physically plausible object layouts following \cite{qin2020s4g}. We use 41000 images for training. We use SPP=128 for ray-tracing rendering with learning-based denoising and sensor noise simulation. 
During training, we use online data augmentation, including color jittering and color drop for RGB images, Gaussian noise and randomly setting depth to zero for depth maps.}

\textbf{{Real-world test dataset.}}
{
In order to collect a real-world test dataset with the ground-truth object detection and pose estimation labels, we first generate scene layouts in the simulation and build the real-world scenes using the sim-real pixel-wise alignment as described in Section~\ref{sec:pixel_wise_alignment}, so that we can acquire precise object poses without the aid of markers.
We collect 504 real images of 24 different scenes using the two depth sensors as the real-world test dataset. }

\textbf{Input modality: depth vs. RGB, RGBD.}
{Here we evaluate the impact of input modality on object detection. We choose YOLOv3 \cite{redmon2018yolov3} as the object detection algorithm. We use three types of input: depth, RGB, and RGBD. We use the pretrained weights on COCO~\cite{lin2014microsoft}, and the weights of the first layer are not loaded for depth and RGBD input. }

\textbf{Experimental results.}
{We use mAP@0.5 (mean of Average Precision with IoU threshold=0.5) as the evaluation metric.
Table~\ref{tab:experiment_object_detection} shows the comparison result of object detection on real-world images. ``Clean depth'' represents using the depth buffer from the renderer. ``Rasterization'' represents simulating the process of active stereovision depth sensor by using rasterization rendering. The performance improvement on RGB images can be attributed to the improved fidelity of ray-tracing rendering, which is consistent with existing works~\cite{zhang2017physically, hodavn2019photorealistic}. Comparing the three input modalities, we can find that depth maps have the best transferability, as they are primarily affected by geometry, whereas RGB images are also affected by surface texture, material, and environmental illumination. It can be shown that, when using the depth map as input, the detection network trained on the simulated data generated by our method has better performance than rasterization and clean depth, demonstrating that our method provides for development of better object detection algorithms.}

\begin{table}[htb]
\caption{Comparison of mAP@0.5 of object detection on real data.\label{tab:experiment_object_detection}}
\centering
\begin{tabular}{@{}c|c|c|c@{}}
\toprule
Data & Depth & RGB & RGBD\\ \hline
Clean depth  & 0.730        & -       & -         \\
Rasterization & 0.942         & 0.839        & 0.875       \\
\textbf{Ours}          & \textbf{0.977}      &  \textbf{0.941}  & \textbf{0.943}       \\ 
\bottomrule
\end{tabular}
\end{table}

\subsection{Generate Training Data for 6D Pose Estimation}
\label{sec: application_pose_estimation}
\textbf{Datasets.}
{We use the same training and test datasets as we use for object detection (Section~\ref{sec:application_object_detection}).}

\textbf{Pose estimation algorithms.}
\label{sec:pose_estimation_algo}
We choose three learning-based 6D pose estimation algorithms:
\begin{enumerate}[(1)]
\item  PVN3D~\cite{he2020pvn3d}. PVN3D predicts the keypoint offsets and semantic segmentation for each point in the point cloud, and uses a deep Hough voting network to predict 6D object poses.

\item  Frustum PointNets~\cite{qi2018frustum}. Frustum PointNets first predict 2D bounding boxes of objects in the depth map and further estimate the 6D object poses using points within the 3D frustums. In our experiments, we use YoloV3~\cite{redmon2018yolov3} for 2D detection and adopt the Hough voting strategy from PVN3D to improve the performance.

\item  SegICP~\cite{wong2017segicp}. SegICP first predicts per-pixel semantic segmentation in the depth map, unprojects the pixels of one object into 3D point cloud, computes the object pose by registering the point cloud with CAD models of different initial poses and selects the result with minimal error as the prediction.

\end{enumerate}

Note that we only use depth information for all the three algorithms, and no RGB information is used.

\textbf{Baseline methods.}
{We compare our proposed depth sensor simulation method with one simple baseline (referred to as \textit{Clean}) that simply uses the depth buffer from the renderer and four state-of-the-art sim-to-real methods: \textit{DepthSynth}~\cite{planche2017depthsynth},  \textit{PixelDA}~\cite{bousmalis2017unsupervised}, Learning to augment (referred to as \textit{LearnAug})~\cite{learningsim2real2019}, and Differentiable Depth Sensor Simulation (referred to as \textit{DDS})~\cite{planche2021physics}}.
{We chose these four works for benchmarking because each represents a distinct type of depth sensor simulation method. DepthSynth simulates the process of an active stereovision depth sensor by rendering binocular images using rasterization rendering and generating the simulated depth by using stereo matching. It represents methods simulating the depth sensor's mechanism.  PixelDA is an unsupervised GAN-based domain adaptation method that decouples the task-specific content and domain appearance, and it represents learning-based domain adaptation methods. LearnAug is a domain randomization method that searches for the optimal sequence of data augmentation policies through Monte-Carlo Tree Search (MCTS) to maximize the trained detection network's performance on a real dataset with ground-truth object detection annotations. It represents domain randomization methods. DDS builds an end-to-end differentiable stereo matching pipeline and optimizes the rendering parameters by using the loss between the real captured depth maps and the generated depth maps. It represents differentiable-rendering-based optimization methods.}

\textbf{{Experimental results.}}
{We evaluate the performance of pose estimation algorithms by using the percentage of predictions whose rotation and translation errors are smaller than certain thresholds. In this paper, we choose two thresholds: (10$\degree$, 10mm), (20$\degree$, 20mm). For symmetrical objects, the rotation error is computed as the smallest error of all possible rotations. In order to analyze the influence of object materials, we further compute the metrics of real objects and 3D-printed objects separately.}

\begin{figure*}[t]
\centering
\includegraphics[width=\textwidth]{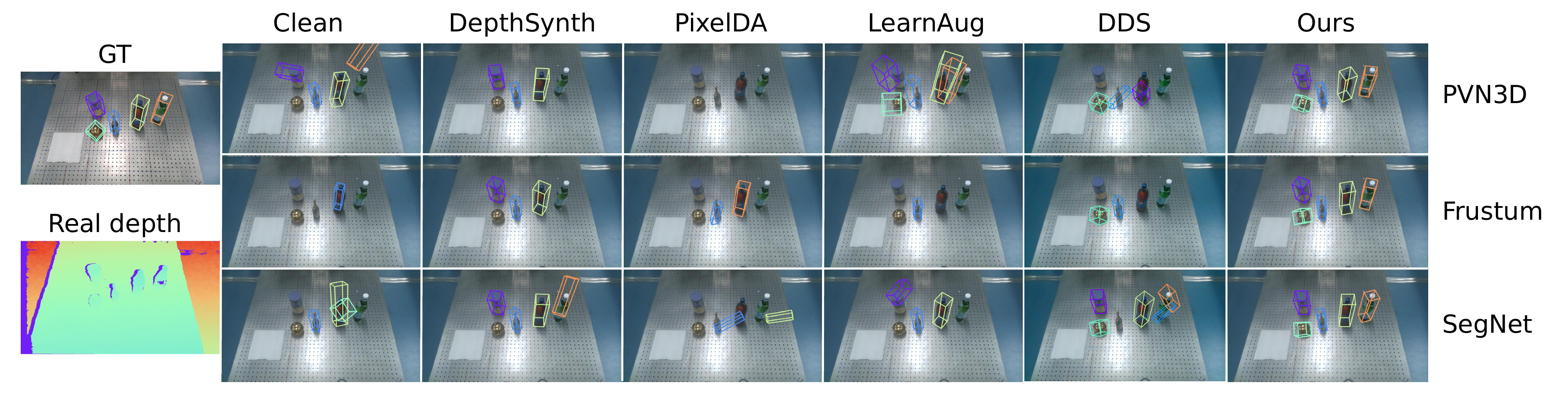}
\caption{Qualitative comparison of 6D object pose estimation algorithms on real depth images. The scene is challenging for pose estimation as the depth measurement of real objects (Golden ball, S.Pellegrino) is noisy and incomplete. All the three pose estimation algorithms are able to infer accurate poses while trained solely on the simulated data generated by our method. Note that we use depth maps for the pose estimation and RGB images are only used for better visualization.}
\label{fig:data_generation}
\end{figure*}
\begin{table*}[h]
\caption{
Comparison of 6D object pose estimation on real depth images. The best result of all three pose estimation algorithms is marked as \textbf{bold}, and the best result of each algorithm is marked as \uline{underlined}. 
\label{tab:data_generation}} 
\centering
\begin{tabular}{@{}c|c|c|c|c|c|c|c@{}}
\toprule
\multirow{2}{*}{Pose algo.} & \multirow{2}{*}{Training data} & \multicolumn{2}{c|}{Overall(\%)} & \multicolumn{2}{c|}{Real(\%)} & \multicolumn{2}{c}{Printed (\%)} \\\cline{3-8}
&&10$\degree$, 10mm  &20$\degree$, 20mm &10$\degree$, 10mm  &20$\degree$, 20mm &10$\degree$, 10mm  &20$\degree$, 20mm\\\hline

\multirow{5}{*}{PVN3D}   & Clean  & 11.54 & 16.47 & 9.59 & 16.62 & 12.82 & 16.03 \\
                         & DepthSynth & 54.06 & 68.93 & 30.49 & 51.52 & 74.09 & 84.07 \\
                         & PixelDA   & 0.00 & 0.00 & 0.00 & 0.00 & 0.00 & 0.00 \\
                         & LearnAug   & 0.32 & 0.48 & 0.35 & 0.71 & 0.28 & 0.28  \\
                         & DDS & 23.57 & 39.21 & 23.11 & 42.42& 23.96 & 36.48\\
                         & \textbf{Ours}  &  \uline{63.37}  & \textbf{83.45} & \uline{42.60} & \textbf{74.81}  & \textbf{80.95}  & \textbf{90.77} \\ \hline
\multirow{5}{*}{Frustum} & Clean     & 2.98 & 4.93 & 1.55 & 2.15 & 4.35 & 7.53 \\
                         & DepthSynth &  40.93 & 68.26 & 34.33 & 62.22 & 46.62 & 74.32 \\
                         & PixelDA   & 15.51 & 26.81 & 12.42 & 27.39 & 18.57 & 27.75 \\
                         & LearnAug   & 18.89 & 30.23 & 11.27 & 18.78 & 26.10 & 41.33 \\
                         & DDS & 20.63 & 35.99 & 21.21& 39.65& 20.15&32.89\\
                         & \textbf{Ours}  & \uline{43.25} & \uline{75.08}          & \uline{38.53 } & \uline{71.08 }         & \uline{47.25}          &\uline{78.46} \\ \hline
\multirow{5}{*}{SegICP}  & Clean     & 5.69 & 7.20 & 2.48 & 3.84 & 8.16 & 9.79 \\
                         & DepthSynth &55.65 & 68.26 & 39.12 & 60.43 & 69.61 & 75.62 \\
                         & PixelDA   & 15.04 & 20.01 & 5.04 & 10.21 & 23.37 & 28.21\\
                         & LearnAug   & 37.07 & 46.50 & 19.37 & 33.17 & 51.70 & 58.16 \\
                         & DDS & 30.71 & 42.30 & 22.16 & 39.13 & 37.95 & 44.98 \\
                         & \textbf{Ours}  & \textbf{65.52}          & \uline{79.64}          &\textbf{50.56} & \uline{74.72}          & \uline{78.17}  & \uline{83.81}\\
\bottomrule
\end{tabular}

\end{table*}

{Fig.~\ref{fig:data_generation} shows the qualitative results. As can be seen in the real depth image, the 3D-printed objects can be properly measured by the real active stereovision depth sensor, whereas the measurement of real objects is noisy and incomplete, which makes it more challenging to simulate. All three pose estimation algorithms can make accurate predictions when trained solely on the simulated data generated by our depth simulator, which validates the effectiveness of our method for generating training data for 6D pose estimation.}

{Table~\ref{tab:data_generation} shows the comparison result of different pose estimation algorithms on real-world depth images. In general, our method can significantly improve the performance of all three algorithms, with the improvement on real objects being greater than that on 3D-printed objects. Because PVN3D uses PointNet++ to extract the geometry information for pose estimation, it is more sensitive to geometry distortion. It has unsatisfactory performances when trained on PixelDA and LearnAug, demonstrating that these two methods introduce additional unrealistic geometric distortions, which is also claimed in \citet{shen2022dcl}. Compared with 3D-printed objects, the performance of DepthSynth on real objects is decreased. It is because 3D-printed objects are Lambertian diffused on which light is uniformly reflected and can be well rendered by rasterization rendering technique, but the light transformation process on real objects is complicated and cannot be correctly rendered by rasterization. The probable reason that DDS's performance is sub-optimal is that it uses a differentiable stereo matching module whose capability is limited compared with SGBM. Moreover, the differentiable rendering used in DDS does not support the rendering of transparent materials.}

\textbf{{Comparison with depth completion methods.}}
{We compare our proposed depth simulation method with two learning-based depth completion methods: ClearGrasp~\cite{sajjan2020clear} and TransCG~\cite{fang2022transcg}. For ClearGrasp, we train the surface normal estimation module on the synthetic RGB images rendered by the ray-tracing renderer, and complete the real depth map using ground-truth boundaries and object masks. For TransCG, we use the pretrained weight\footnote{\href{https://github.com/galaxies99/transcg}{https://github.com/galaxies99/transcg}} released by the authors to complete the depth map.
For these two methods, we train the pose estimation algorithms on the synthetic \textit{clean} depth and evaluate them on the depth completion outputs. Table~\ref{tab:depth_completion} shows the comparison results. For all three algorithms, the pose estimation accuracies on completed depth maps are worse than those on the original captured depth maps. This is likely due to the limited generalizability of learning-based depth completion methods to unseen objects and scenes, which may introduce more artifacts to depth maps and reduce pose estimation algorithms' transferability.}

\begin{table}[htb]
\caption{
{Comparison of depth completion and depth sensor simulation.}
\label{tab:depth_completion}} 
\centering
\begin{tabular}{@{}c|c|c|c|c@{}}
\toprule

\multirow{2}{*}{Pose algo.}  &\multirow{2}{*}{\makecell[c]{Training\\ data}} & \multirow{2}{*}{\makecell[c]{Test\\ data}} & \multicolumn{2}{c}{Overall(\%)} \\\cline{4-5}
& & &{10$\degree$, 10mm}  &{20$\degree$, 20mm} \\\hline
\multirow{4}{*}{PVN3D} & Clean & Real & 13.73 &  19.56 \\
 & Clean & ClearGrasp& 1.67 & 6.43\\
& Clean & TransCG& 0.00 & 0.04 \\\cline{2-5}
& Ours & Real & \textbf{63.37} & \textbf{83.45}\\\hline
\multirow{4}{*}{Frustum}  & Clean & Real  & 3.85   &  5.12\\
 &Clean & ClearGrasp & 0.16 & 0.52\\
&Clean & TransCG & 0.00 & 0.12 \\ \cline{2-5}
&Ours & Real & \textbf{43.25} & \textbf{75.08}\\\hline
\multirow{4}{*}{SegICP}  & Clean & Real & 5.83 & 7.46\\
 & Clean & ClearGrasp & 1.39 & 3.29\\
  &Clean & TransCG & 0.48 & 0.71 \\ \cline{2-5}
        & Ours & Real& \textbf{65.52} & \textbf{79.64}\\

\bottomrule
\end{tabular}

\end{table}

\subsection{Generate Training Data for Robot Grasping}
\label{sec:application_robot_grasping}

\textbf{Experimental setting.}
In this section, we choose robot grasping as the task to validate the effectiveness of our method for continuous policy learning, as it is one of the most fundamental robotic tasks. The experimental setting is similar to~\cite{viereck2017learning} and~\cite{doi:10.1177/0278364919859066}. To evaluate the sim-to-real performance of different objects, we only focus on the grasping of a single object. The input is the depth map from the wrist-mounted depth sensor (3B in Fig.~\ref{fig:experiment_setup}). The action of each step  is $a=[\Delta x_g, \Delta y_g, \Delta z_g, \Delta \theta_g]\in \mathbb{R}^4$, where $(x_g, y_g, z_g)$ is the gripper position, $\theta_g$ is the rotation of the gripper around the $z$-axis. After $N_g$ steps, the gripper goes down a predefined distance, closes, and lifts. The grasp is considered as success if the object is lifted stably.

{We follow the pipeline of~\cite{mu2021maniskill} to train the grasping policy: we first train an RL policy using privileged states as input, and train a depth-map-based policy by using behavior cloning (BC). We directly deploy the trained depth-map-based policy on the real robot without any finetuning.}

\textbf{Experimental details.}
We choose 6 optically challenging real objects and 2 3D-printed objects for the real grasping experiment. The object is randomly placed on the table with the range of $x$ 0.15m, $y$ 0.30m, $\theta$ 0.6rad. $N_g=10$.
For the RL part, we choose SAC~\cite{haarnoja2018soft} as the RL algorithm. {We use Adam as the optimizer with a learning rate of $1\times 10^{-3}$, buffer size $10^6$, discount coefficient 0.99 and target smoothing coefficient 0.01.} The agent is trained for 540k steps and evaluated for every 8k steps. The network weights of the best performance are chosen for the following BC part.

For the BC part, the resolution of the depth map is $352\times224$. The number of demonstrations is $~6\times 10^{5}$. Data augmentation, including Gaussian noise and randomly setting depth to zero, is used for both our method and baseline methods. { We use Adam as the optimizer with a learning rate of $1\times10^{-4}$, batch size 256, and 100 epochs.} The model of the best performance in simulation is used for grasping in the real world. {We use SPP=8 with learning-based denoiser for a shorter rendering time cost.} 

\begin{figure}[htb]
    \centering
    \includegraphics[width=0.5\linewidth]{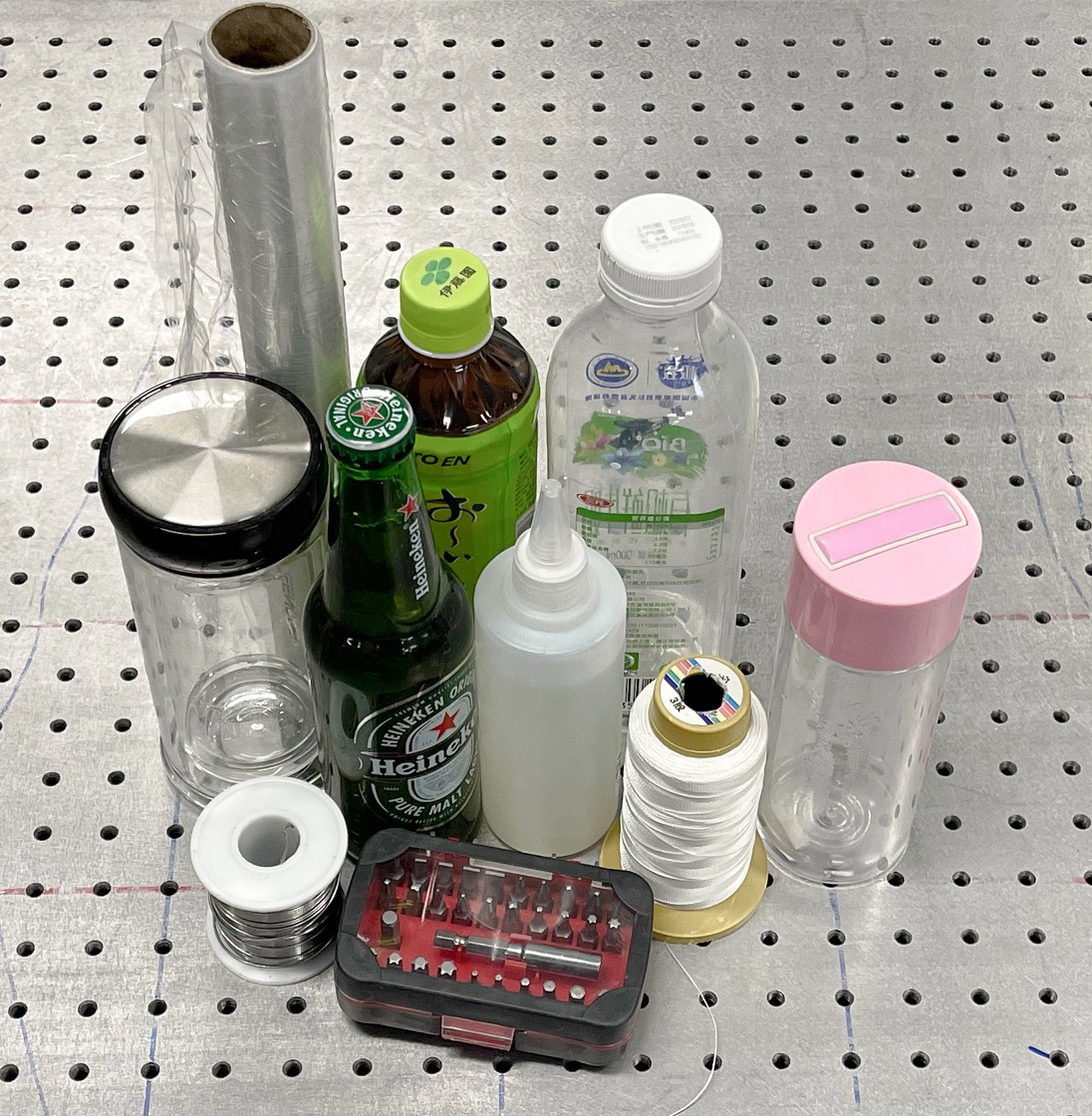}
    \caption{10 unseen optically challenging objects for evaluating the generalizability of the grasping policy.}
    \label{fig:grasp_generalize}
\end{figure}

\textbf{Experimental results.}
For the real grasping experiment, each object is randomly placed on the table 10 times within the same range as simulation training.
Table~\ref{tab:rl_grasp} shows the comparison results of grasping in the real world for continuous policy learning on different simulated depths. {The policy trained on our simulated depth achieves an overall success rate of 96.25\%, which validates the effectiveness of our method for continuous policy learning algorithm training.}
Note that the success rate of grasping different objects is dependent not only on the sim-to-real transferability of the policy but also on the target object's shape and friction coefficient. For instance, Golden ball cannot be grasped if there is a little offset between the gripper and the ball, whereas Voss and Sharpener may be grasped with a larger offset.
We further evaluate the trained grasping policy on 10 unseen optically challenging objects as shown in Fig.~\ref{fig:grasp_generalize}. It achieved a success rate of 100\%, which validates the generalizability of the trained policy.

\begin{table}[htb]
\caption{
Experimental results of grasping in real world for continuous policy learning on different simulated depths.
\label{tab:rl_grasp}} 
\centering
\begin{tabular}{@{}c|c|c|c|c@{}}
\toprule
\multirow{2}{*}{\makecell[c]{Object\\ type}} & \multirow{2}{*}{\makecell[c]{Object\\ name}} & \multicolumn{3}{c}{Simulated Depth} \\\cline{3-5}
                             &                              & Clean   & DepthSynth   & Ours  \\\hline
\multirow{6}{*}{Real}        & Coca Cola                    &    7/10         &    9/10          & \textbf{ 10/10}      \\
                             & Jack Daniels                 &     4/10        &    7/10          &   \textbf{8/10}    \\
                             & Golden ball                       &     3/10        &    6/10          &    \textbf{10/10}   \\
                             & S.Pellegrino                 &    3/10         &     3/10         & \textbf{7/10}      \\
                             & Tennis ball                  &   \textbf{10/10}          &   7/10           &  \textbf{ 10/10}    \\
                             & Voss                         &     8/10        &      8/10        &  \textbf{10/10}     \\ \midrule
\multirow{2}{*}{Printed}     & Beer Can                     &     5/10        &   8/10           &   \textbf{10/10}    \\
                             & Sharpener                    &     9/10        &    7/10          &  \textbf{10/10}    \\ \midrule
                             
\multicolumn{2}{c|}{Overall} & 49/80 & 55/80 & \textbf{77/80}\\ \bottomrule
                             
\end{tabular}

\end{table}
\subsection{Estimate the Ranking of Algorithms in Real World}
\label{sec:application_algorithm_benchmarking}
\textbf{Experimental setting.}
{In this section, we validate the effectiveness of our method as algorithm benchmarking for 6D object pose estimation algorithms. We choose the three pose estimation algorithms: PVN3D, Frustum, and SegICP. For each algorithm, we train it on three simulated depth image datasets generated by different methods (Clean, DepthSynth, ours) and make $3\times3=9$ pose estimation models with different performances for algorithm benchmarking. We evaluate the performances of the 9 models on both the real test data and the simulated test data generated using baseline methods and our method, respectively. Because the real depth images and the simulated depth images are precisely aligned as described in Section~\ref{sec:pixel_wise_alignment}, the evaluation results on the real-world test dataset and the simulated test dataset can be compared directly.}

\textbf{Experimental results.}
{An ideal simulated test dataset should deliver evaluation results of different models that are consistent with real-world evaluation results. Therefore, we compute the correlation coefficients of the evaluation results between the real-world test dataset and the simulated test dataset. Table~\ref{tab:algorithm_benchmarking} shows the experimental results. Our method can deliver more consistent evaluation results than baseline methods on both metrics. In addition, the difference between our method and DepthSynth is greater on real objects than that on 3D-printed objects, which demonstrates that our method can simulate the depth error pattern on optically challenging objects more accurately than DepthSynth. }
\begin{figure}[htb]
\centering
\subfloat[]{\includegraphics[width=0.49\linewidth]{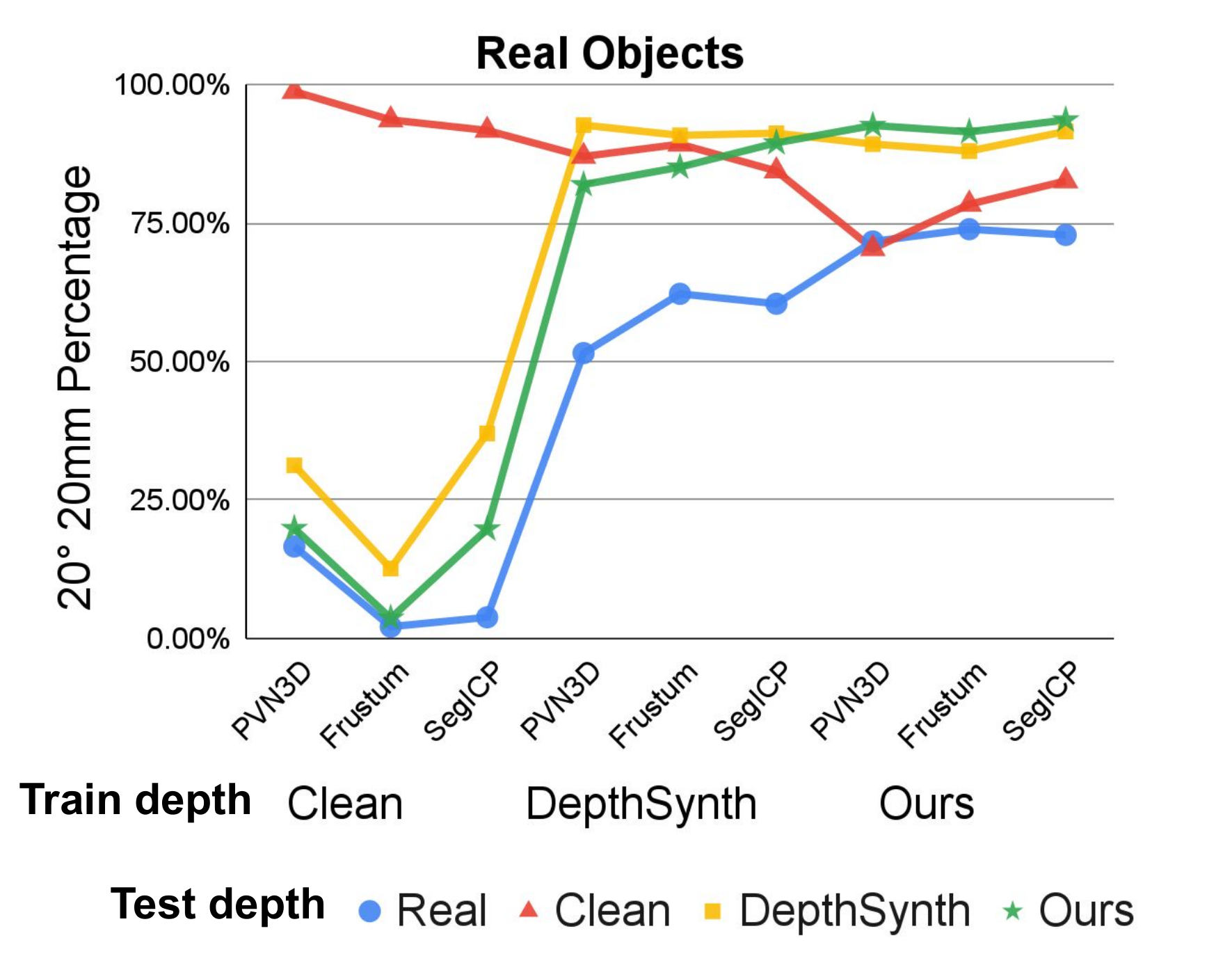}%
}
\hfil
\subfloat[]{\includegraphics[width=0.49\linewidth]{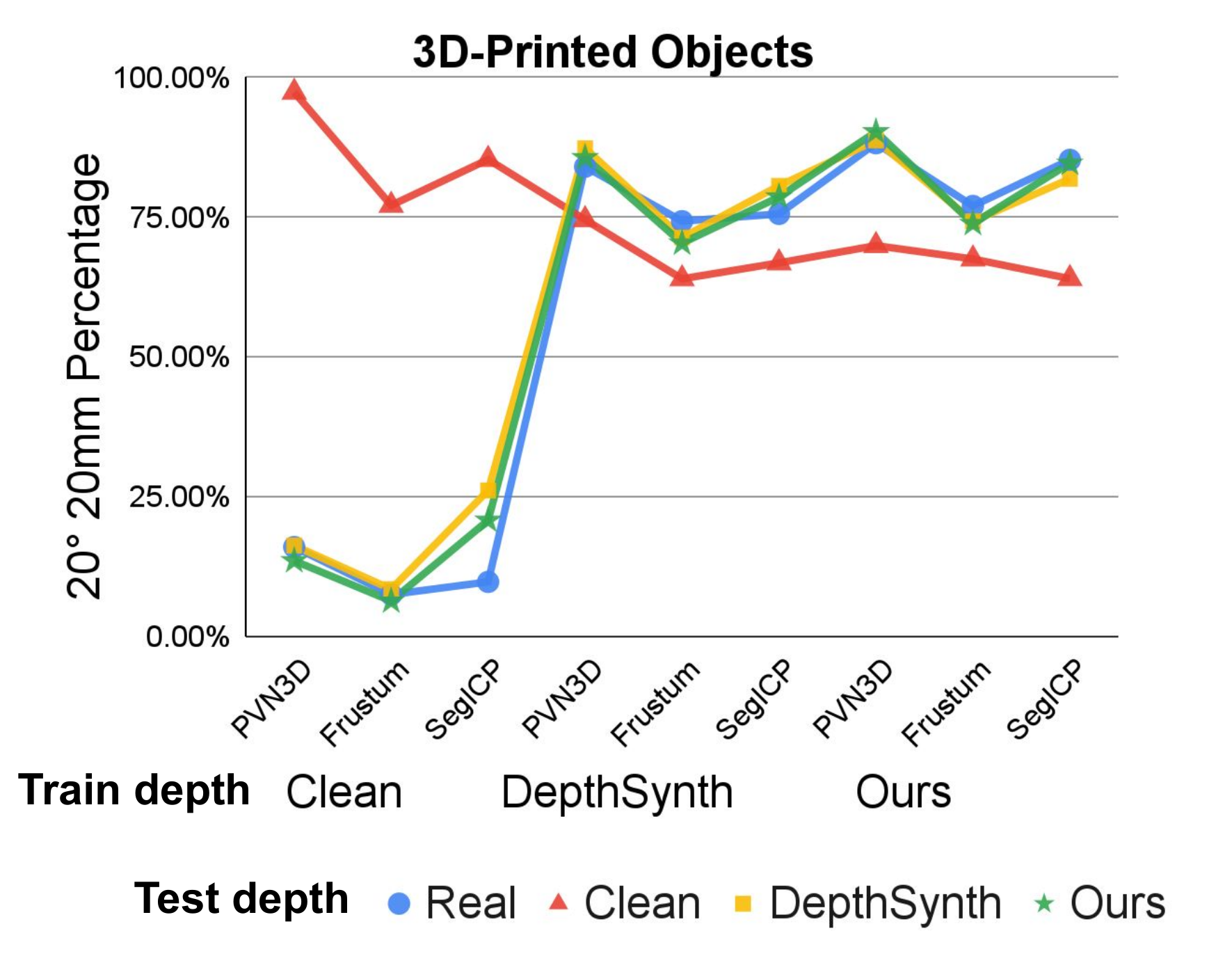}%
}
\caption{Metric curve of 9 pose estimation models on (a) real objects and (b) 3D-printed objects.}
\label{fig:algorithm_benchmarking_curve}
\end{figure}

\begin{figure*}[htb]
\centering
\includegraphics[width=0.8\textwidth]{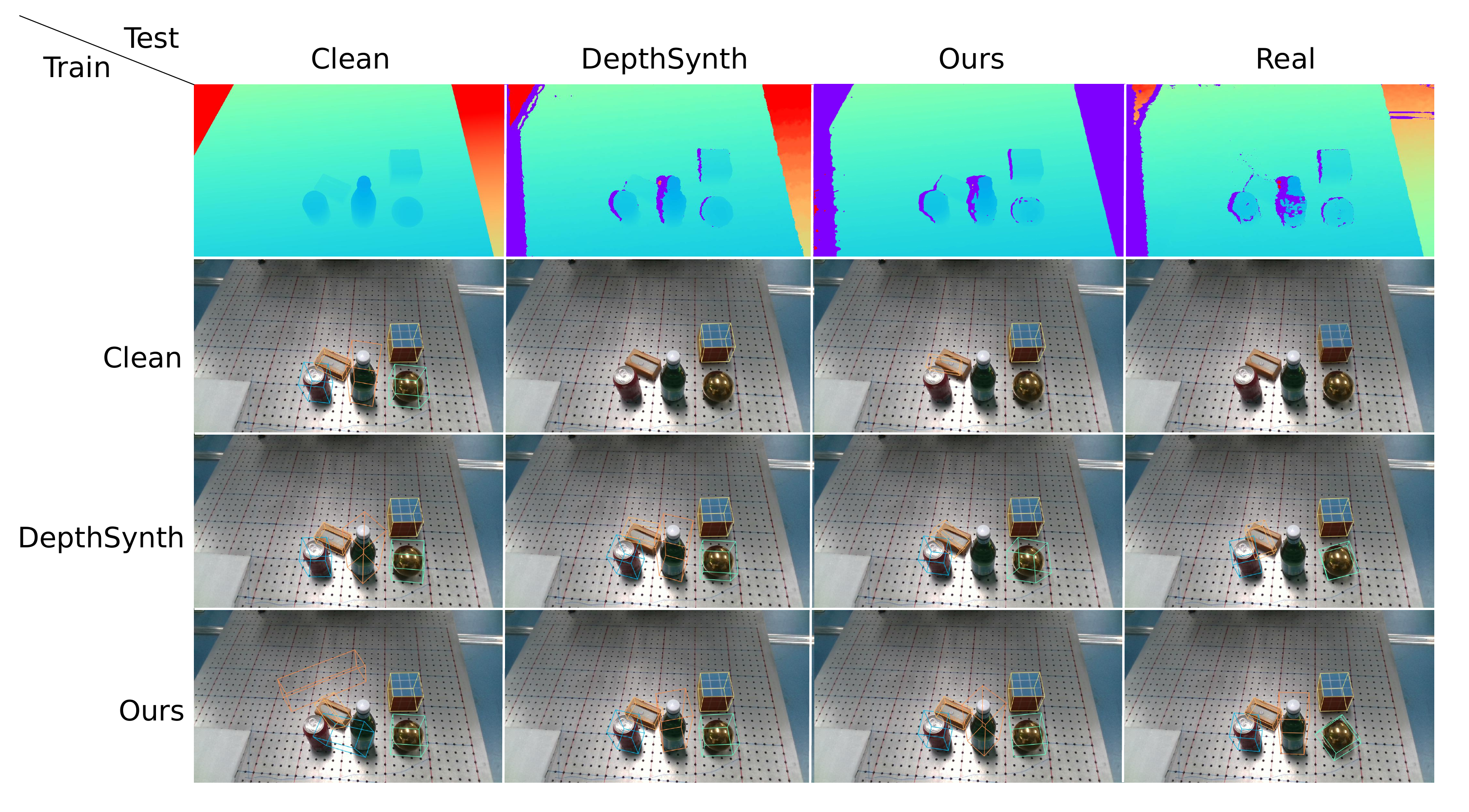}
\caption{Qualitative comparison of depth images of different simulation methods and real sensor and the pose estimation results of PVN3D trained and tested on different depth data. Our method can simulate the depth noise pattern on optically challenging objects (specular, transparent). The evaluation results on simulated test data generated by our method are consistent with the real evaluation results for all the three models trained with different simulated data.}
\label{fig:algo_benchmarking}
\end{figure*}
{Fig.~\ref{fig:algorithm_benchmarking_curve} shows the performance of the 9 models on real objects and 3D-printed objects. The evaluation results of our method are aligned with the results on real-world depth maps for both kinds of objects. For real objects, the evaluation results of models trained on DepthSynth have higher scores when tested on DepthSynth, which is inconsistent with the results on real-world depth maps. Fig.~\ref{fig:algo_benchmarking} shows the depth maps of different simulation methods and the real sensor and the 6D object pose estimation results of PVN3D trained on different simulated data and tested on different simulated data along with real data. Our method can simulate the holes and noise on the specular and transparent objects as the depth map captured by the real depth sensor. For the models trained on different simulated data, the evaluation results on our simulated depth are aligned with the ones on real-world depth maps. For the model trained on DepthSynth, it can estimate the object poses accurately from DepthSynth simulated depth, but fail on both our simulated depth and the real-world depth.}

\begin{table*}[htb]
\caption{
Correlation coefficients of 6D object pose estimation results between different test depth images and real test depth images. 
\label{tab:algorithm_benchmarking}} 
\centering
\begin{tabular}{@{}c|c|c|c|c|c|c@{}}
\toprule
\multirow{2}{*}{\makecell[c]{Test\\Depth}} & \multicolumn{2}{c|}{Overall} & \multicolumn{2}{c|}{Real} & \multicolumn{2}{c}{Printed } \\ \cline{2-7}
&10$\degree$, 10mm  &20$\degree$, 20mm &10$\degree$, 10mm  &20$\degree$, 20mm &10$\degree$, 10mm  &20$\degree$, 20mm\\\hline
Clean  & -0.562 & -0.867 & -0.689 & -0.791 & -0.383 & -0.790 \\
DepthSynth & 0.973 & 0.977   & 0.925  & 0.947 & 0.988   & 0.987  \\ 
Ours       & \textbf{0.982} & \textbf{0.991}  & \textbf{0.959} & \textbf{0.986} & \textbf{0.993}  & \textbf{0.992} \\
\bottomrule
\end{tabular}

\end{table*}

\section{Ablation study}
\label{sec:ablation_study}

In this section, we conduct experiments to validate our design choices and evaluate the influence of key hyperparameters. We choose the application of generating training data for 6D pose estimation for the ablation study. According to the order of our sensor simulation pipeline, we first validate the effectiveness of material acquisition (Section~\ref{sec:ablation_material}), then the effect of rendering settings (Section~\ref{sec:ablation_rendering}), and finally the sensor noise simulation (Section~\ref{sec:ablation_noise}).

\subsection{{Material acquisition}}
\label{sec:ablation_material}

{Table~\ref{tab:ablation_material} shows the comparison result of different pose estimation algorithms trained on different material parameters. Our method is compared to a default material parameter set and three random material parameter sets. The default material is set to roughness=0, metallic=0, specular=0,  transmission=1 for Jack Daniels, S.Pellegrino and VOSS, and transmission=0 for all other objects. The performance on real objects are consistently improved for all the three algorithms, which demonstrates that the simulation of optically challenging objects requires more accurate material parameters.  By combining plain $L_2$ loss and the perceptual loss, we can improve the performance across the majority of metrics, because $L_{percept}$ is more robust to exposure and lighting condition misalignment.} 
\begin{table*}[htb]
\caption{
{Ablation study on material acquisition for generating training data for 6D pose estimation. The best result of all three pose estimation algorithms is marked as \textbf{bold}, and the better result of each algorithm is marked as \uline{underlined}.}
\label{tab:ablation_material}} 
\centering
\begin{tabular}{@{}c|c|c|c|c|c|c|c@{}}
\toprule
\multirow{2}{*}{Pose algo.} & \multirow{2}{*}{\makecell[c]{Material\\parameters}} & \multicolumn{2}{c|}{Overall(\%)} & \multicolumn{2}{c|}{Real(\%)} & \multicolumn{2}{c}{Printed (\%)} \\\cline{3-8}
&&10$\degree$, 10mm  &20$\degree$, 20mm &10$\degree$, 10mm  &20$\degree$, 20mm &10$\degree$, 10mm  &20$\degree$, 20mm\\\hline
\multirow{6}{*}{PVN3D}      & Default                   & 60.87          & 79.44          & 37.23         & 66.58        & 80.88          & 90.33          \\
                            & Random1                   & 62.70          & 80.28          & 42.16         & 70.13        & 80.07          & 88.86          \\
                            & Random2                   & 62.66          & 80.75          & 40.00         & 67.10        & 81.83          & \textbf{92.31}          \\
                            & Random3                   & 62.10          & 79.88          & 40.09         & 67.45        & 80.73          & 90.40          \\\cline{2-8}
                            & Ours w/o  $L_{percept}$ & \textbf{64.09}          & 83.02          & \uline{43.12}         & 72.73        & \textbf{81.83}          & 91.72          \\ 
                            & Ours                   & 63.29          & \textbf{83.25}          & 42.60         & \textbf{74.63}        & 80.81          & 90.55          \\ \hline
\multirow{6}{*}{Frustum}    & Default                   & 41.67          & 73.53          & 34.63         & 66.75        & \uline{47.62}          & \uline{79.27}          \\
                            & Random1                   & \uline{43.57}          & \uline{74.48}          & 39.05         & 70.82        & 47.40          & 77.58          \\
                            & Random2                   & 42.18          & 73.89          & 36.88         & 67.97        & 46.67          & 78.90          \\
                            & Random3                   & 40.99          & 71.83          & 37.40         & 68.31        & 44.03          & 74.80          \\\cline{2-8}
                            & Ours w/o  $L_{percept}$ & 42.62          & 72.66          & 36.80         & 67.53        & 47.55          & 77.00          \\
                            & Ours                        & 42.42          & 74.37          & \uline{39.39}         & \uline{72.38}        & 44.98          & 76.04          \\ \hline
\multirow{6}{*}{SegICP}     & Default                   & 55.04          & 66.55          & 39.13        & 58.96        & 68.50         & 72.97        \\
                            & Random1                   & 42.18          & 49.64          & 36.45         & 50.22        & 47.03          & 49.16          \\
                            & Random2                   & 62.38          & \uline{76.55}          & 44.94         & 68.92        & \uline{77.14}          & \uline{83.00}          \\
                            & Random3                   & 56.27          & 68.06          & 42.86         & 62.51        & 67.62          & 72.75          \\\cline{2-8}
                            & Ours w/o  $L_{percept}$ & 50.63          & 60.28          & 36.28         & 53.59        & 62.78          & 65.93          \\
                            & Ours                       & \uline{62.66}          & 76.43          & \textbf{46.93}         & \uline{71.17}        & 75.97          & 80.88         \\
\bottomrule
\end{tabular}

\end{table*}
\subsection{{Rendering settings}}
\label{sec:ablation_rendering}
\begin{figure}
\centering
\subfloat[]{\includegraphics[width=0.49\linewidth]{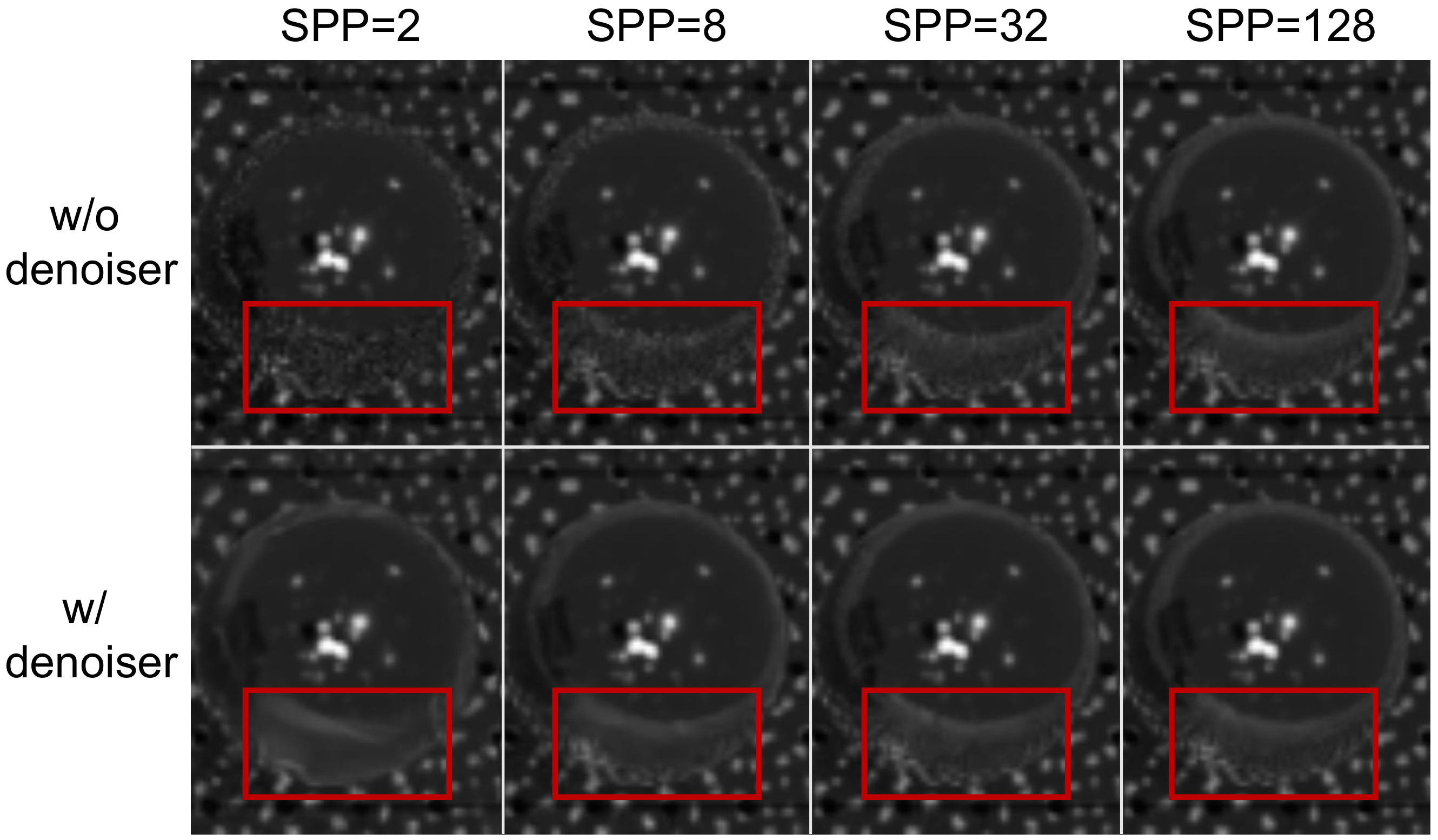}%
}
\hfil
\subfloat[]{\includegraphics[width=0.49\linewidth]{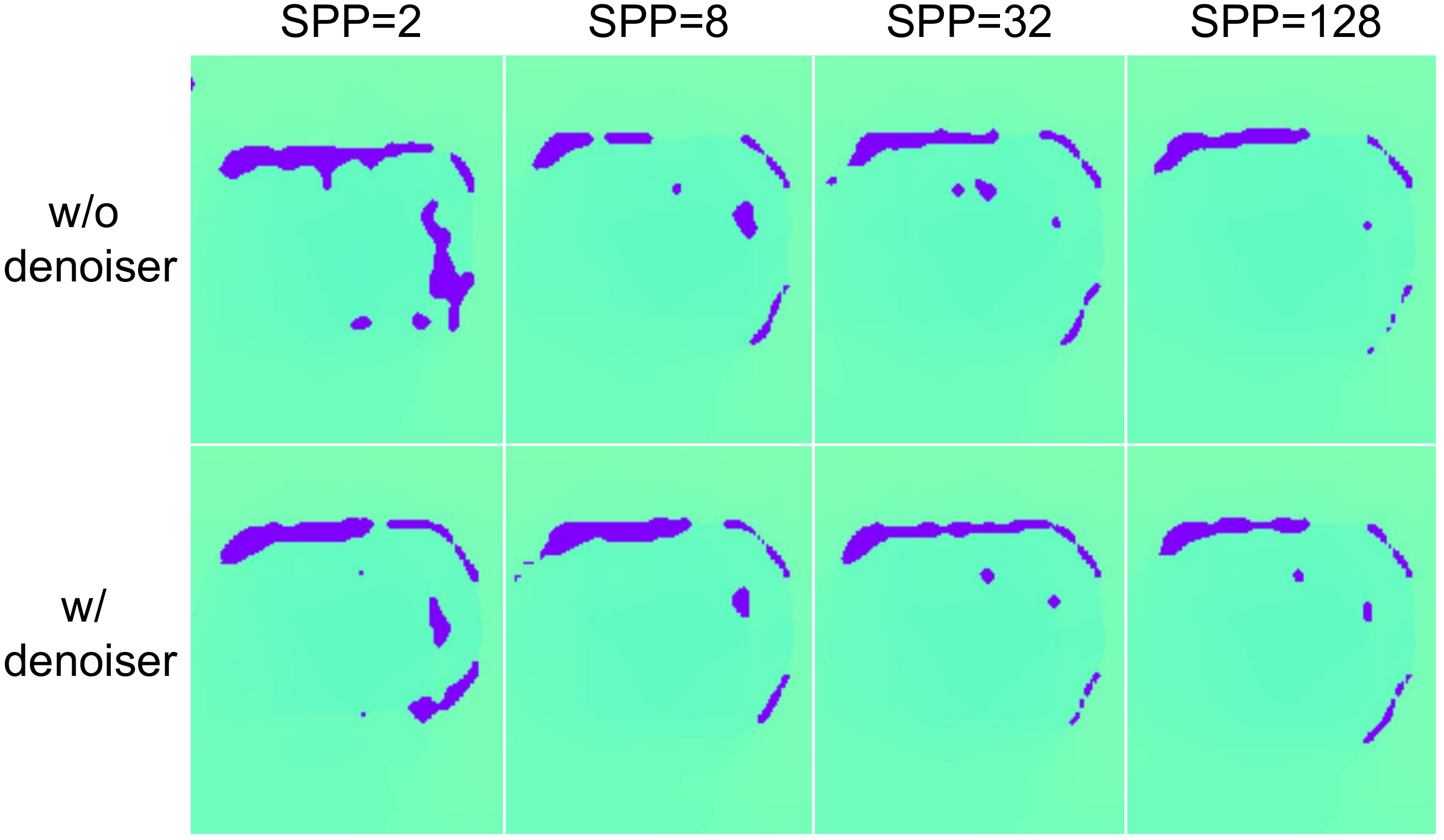}%
}
\caption{{(a) Rendered IR image and (b) simulated depth maps of different rendering settings.}}
\label{fig:spp_ablation}
\end{figure}

{Table~\ref{tab:ablation_rendering} shows the comparison result of different rendering settings: SPP and whether to use the learning-based denoiser. We set SPP to 2, 8, 32, 128.  It can be seen that using the denoiser improves the performance across the majority of metrics and the impact of SPP is not remarkable for all the three algorithms. As shown in Fig.~\ref{fig:spp_ablation}, when SPP is low, the IR pattern on the metallic ball is noisy and is erased by the learning-based denoiser. Since the two rendered IR images are equally affected by the denoiser and SPP, the effect is decreased for stereo matching depth maps. Therefore, when focusing on the transferability of depth maps, we can use a low SPP with the denoiser to simulate the depth, whereas we better use a high SPP when aiming for high-fidelity RGB or IR images.}

\begin{table*}[htb]
\caption{
{Ablation study on rendering settings, including SPP and denoising. The best result of all three pose estimation algorithms is marked as \textbf{bold}, and the best result of each algorithm is marked as \uline{underlined}.} 
\label{tab:ablation_rendering}} 
\centering
\begin{tabular}{@{}c|c|c|c|c|c|c|c|c@{}}
\toprule
\multirow{2}{*}{Pose algo.} & \multirow{2}{*}{SPP} & \multirow{2}{*}{Denoiser}  & \multicolumn{2}{c|}{Overall(\%)} & \multicolumn{2}{c|}{Real(\%)} & \multicolumn{2}{c}{Printed (\%)} \\\cline{4-9}
&&&10$\degree$, 10mm  &20$\degree$, 20mm &10$\degree$, 10mm  &20$\degree$, 20mm &10$\degree$, 10mm  &20$\degree$, 20mm\\\hline
\multirow{8}{*}{PVN3D}   & 2   & \ding{55} &   64.33          & 83.85          &\uline{44.94} & 75.58          & 80.73          & 90.84          \\
                         & 2   & \checkmark  &  \uline{65.00} &\textbf{85.91} & 43.55          &\textbf{77.58} &\textbf{83.15} &\textbf{92.97} \\
                         & 8   & \ding{55} &   60.48          & 79.88          & 41.39          & 72.38          & 76.63          & 86.23          \\
                         & 8   & \checkmark   & 62.42          & 83.25          & 41.99          & 74.72          & 79.71          & 90.48          \\
                         & 32  & \ding{55} &   63.89          & 84.44          & 42.42          & 75.67          & 82.05          & 91.87          \\
                         & 32  & \checkmark  &  63.29          & 83.25          & 42.60          & 74.63          & 80.81          & 90.55          \\
                         & 128 & \ding{55} &  63.29          & 82.46          & 43.72          & 73.16          & 79.85          & 90.33          \\
                         & 128 & \checkmark  &  63.37          & 83.45          & 42.60          & 74.81          & 80.95          & 90.77          \\ \hline
\multirow{8}{*}{Frustum} & 2   & \ding{55} &  42.42          & 74.60          & 38.87          &\uline{73.77} & 45.42          & 75.31          \\
                         & 2   & \checkmark  &  42.42          & 75.36          & 36.19          & 72.03          & 47.69          & 78.17          \\
                         & 8   & \ding{55} &  41.90          & 75.12          & 36.71          & 72.38          & 46.30          & 77.44          \\
                         & 8   & \checkmark  &  43.10          &\uline{75.63} & 37.58          & 72.38          &\uline{47.77} & 78.39          \\
                         & 32  & \ding{55} &  42.06          & 74.76          & 37.14          & 71.00          & 46.23          & 77.95          \\
                         & 32  & \checkmark  &  42.42          & 74.37          &\uline{39.39} & 72.38          & 44.98          & 76.04          \\
                         & 128 & \ding{55} &  42.70          & 74.96          & 38.35          & 71.60          & 46.37          & 77.80          \\
                         & 128 & \checkmark  & \uline{43.25} & 75.08          & 38.53          & 71.08          & 47.25          &\uline{78.46} \\ \hline
\multirow{8}{*}{SegICP}  & 2   & \ding{55} &  64.92          & 80.60          & 47.88          & 74.55          & 79.34          &\uline{85.71} \\
                         & 2   & \checkmark  &  65.44          & 81.19          & 48.40          & 76.19          &\uline{79.85} & 85.42          \\
                         & 8   & \ding{55} & \textbf{66.23} &\uline{81.51} & 50.48          &\uline{76.54} & 79.56          &\uline{85.71} \\
                         & 8   & \checkmark  &  65.40          & 80.79          & 48.48          & 74.98          & 79.71          &\uline{85.71} \\
                         & 32  & \ding{55} &  50.04          & 60.63          & 38.53          & 58.27          & 59.78          & 62.64          \\
                         & 32  & \checkmark  &  62.66          & 76.43          & 46.93          & 71.17          & 75.97          & 80.88          \\
                         & 128 & \ding{55} &  52.62          & 62.34          & 40.61          & 56.80          & 62.78          & 67.03          \\
                         & 128 & \checkmark  &  65.52          & 79.64          &\textbf{50.56} & 74.72          & 78.17          & 83.81          \\ 

\bottomrule
\end{tabular}
\end{table*}

\subsection{{Sensor noise simulation}}
\label{sec:ablation_noise}
{
Table~\ref{tab:ablation_sensor_noise} shows the comparison of different sensor noise scales. For the majority of metrics, the best performance is observed with the identified sensor scale. When using a more aggressive noise, the trained algorithm may become more robust to noise, but performance is compromised as a result.}

\begin{table}[htb]
\caption{
{Ablation study on sensor noise simulation. The best result of all three pose estimation algorithms is marked as \textbf{bold}, and the better result of each algorithm is marked as \uline{underlined}}.
\label{tab:ablation_sensor_noise}} 
\centering
\begin{tabular}{@{}c|c|c|c@{}}
\toprule
\multirow{2}{*}{Pose algo.}  & \multirow{2}{*}{{Noise Scale}} & \multicolumn{2}{c}{Overall(\%)} \\\cline{3-4}
&&{10$\degree$, 10mm}  &{20$\degree$, 20mm} \\\hline
\multirow{6}{*}{PVN3D} & 0.0  & 61.71 & 79.25 \\
 & 0.1 & 62.62 & 81.11\\
  & 0.3& 62.30 & 81.63\\
& 1.0 & \uline{63.37}  & \textbf{83.45} \\ 
& 3.0 &60.71  & 80.71 \\ 
& 10.0 & 46.19 &61.11 \\\hline
\multirow{6}{*}{Frustum}  & 0.0  & 41.31 & 72.42 \\
 & 0.1 & 42.66 & \uline{75.12}\\
  & 0.3 & 42.38 & 73.61\\
& 1.0 & \uline{43.25} & 75.08 \\ 
& 3.0 & 42.74 & 74.68 \\ 
& 10.0 &36.98 & 60.56\\\hline
\multirow{6}{*}{SegICP}  & 0.0    & 63.29 & 77.02\\
 & 0.1 & 65.12 & 78.97\\
  & 0.3 & \textbf{65.56} &79.21\\
    & 1.0 & 65.35 & \uline{79.72} \\ 
    & 3.0 & 62.74 & 77.02 \\ 
    & 10.0 & 21.59 & 28.81\\

\bottomrule
\end{tabular}

\end{table}

\section{Conclusion and future work}
\label{sec:conclusion}

Despite the thrive of novel algorithms developed in simulators, there has always been strong dispute on whether algorithms learned in simulators can transfer to the real world, among engineers, paper reviewers, and conference attendees. The opponents of robot simulators argue that, obtaining high-quality simulation, if possible at all, will be extremely costly in modeling the 3D environment and synthesizing realistic data. 

\begin{figure}
    \centering
    \includegraphics[width=\linewidth]{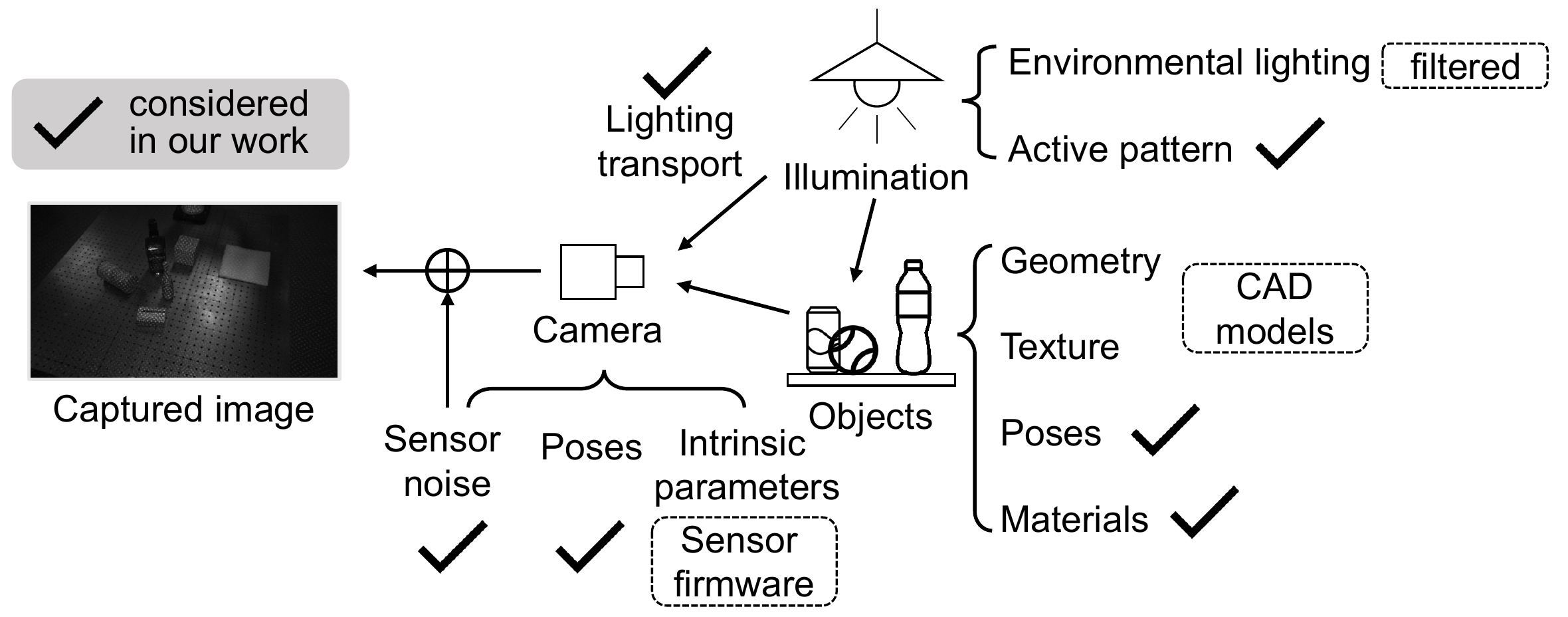}
    \caption{The image capturing process in the real world{.}}
    \label{fig:capture_framework}
\end{figure}

In this paper, we have presented a physics-grounded active stereovision depth sensor simulator that can generate low optical sensing domain gap depth maps with realistic material-dependent error patterns.
As validated by the experimental results, our depth sensor simulator can significantly narrow the sim-to-real optical sensing domain gap and enable the trained algorithms to transfer to the real world without any fine-tuning. We attribute this achievement to the reasons as follows.

{As shown in Fig.~\ref{fig:capture_framework}, images captured by CMOS in the real world are determined by the camera, the illumination and the objects. Our work takes into account most of these factors in a physics-grounded fashion.}
Most importantly, by simulating the active stereovision depth sensor instead of passive RGB cameras, we eliminate the complex influence of environmental lighting in the real world and replace it with actively projected pattern that is known and controllable. In fact, we believe that we leverage a key advantage of robot systems which deserve the exploration by many future works -- compared with humans, robots can customize their sensing system and actively emit controllable signals to facilitate sensing. By taking this advantage, we can simplify the simulation process and better narrow the sim-to-real domain gap. We look forward to many future works to dig deeper in leveraging this advantage in the sim-to-real transfer field.

In the future, we would further close the sim-to-real gap of depth sensing by modeling more advanced effects \textit{e.g.} motion blur and rolling shutter effect. Also, as revealed in this work, object material capturing is a key to credible sensor simulation. Our current object material acquisition method requires the geometry to be precisely aligned between the simulation and the real world, which is more difficult and time-consuming for deformable objects. For future work, we plan to develop a novel material acquisition method that supports unpaired simulated and real images. Moreover, our current method does not account for the fact that some material properties of real deformable objects can change due to interaction. We will also study and integrate more type of sensors in our simulation platform to facilitate more sim-to-real studies in robotics and vision community.

\bibliographystyle{IEEEtranN}
\bibliography{references}
\vspace{-16mm}
\begin{IEEEbiography}[{\includegraphics[width=1in,height=1.25in,clip,keepaspectratio]{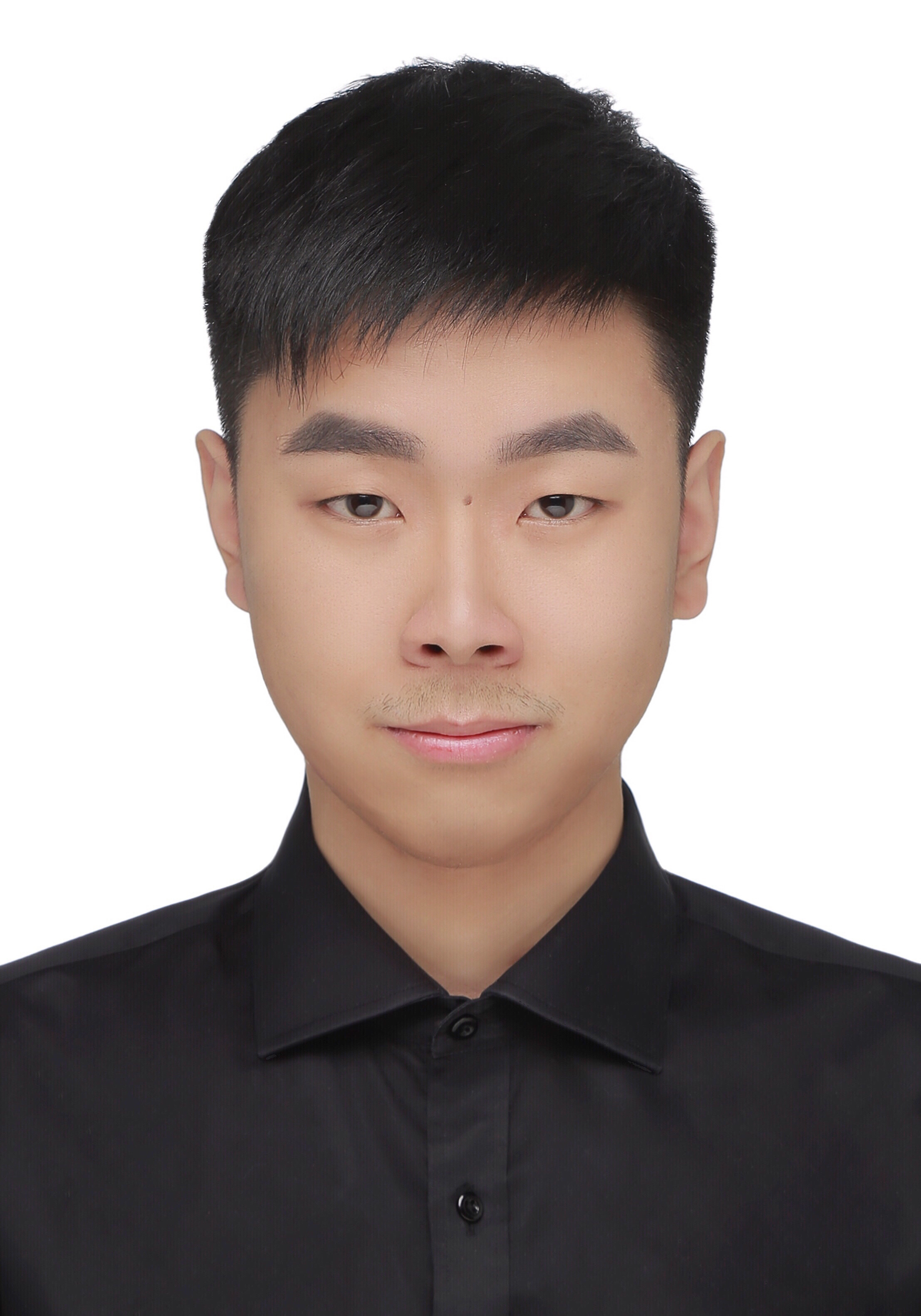}}]{Xiaoshuai Zhang} is a Ph.D. student in Computer Science at the University of California, San Diego. He received the B.S. degree from Peking University, Beijing, China in 2019. His current research interests include 3D scene reconstruction and understanding, computer graphics and simulation.
\end{IEEEbiography}
\vspace{-12mm}
\begin{IEEEbiography}[{\includegraphics[width=1in,height=1.25in,clip,keepaspectratio]{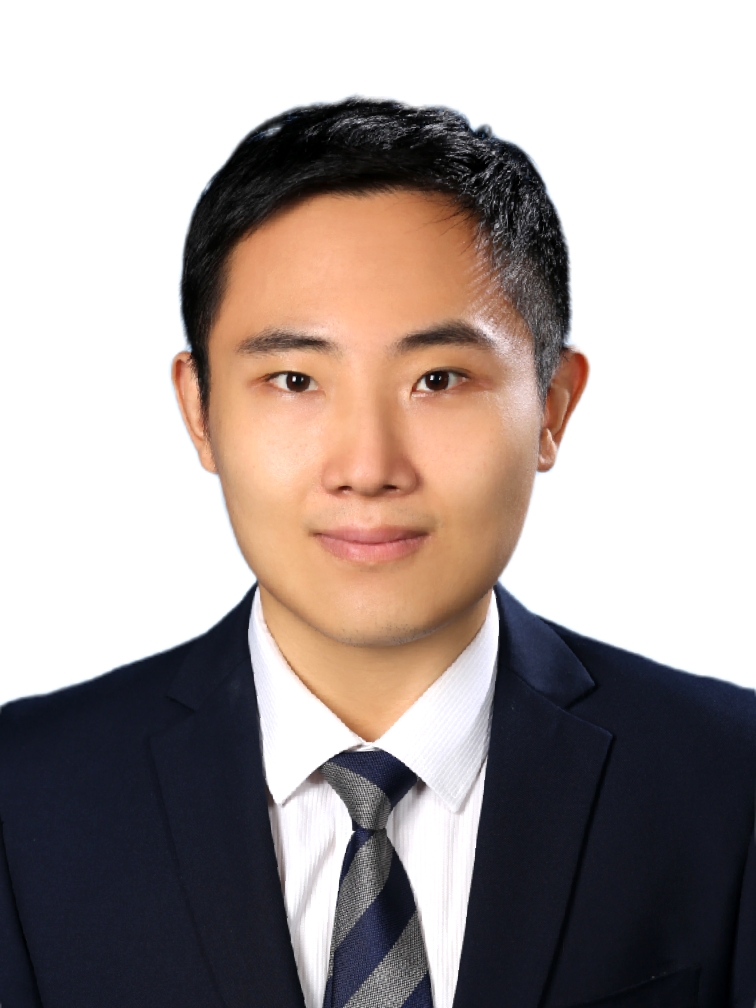}}]{Rui Chen} is currently a research assistant professor in the Department of Mechanical Engineering, Tsinghua University. He received the Ph.D. degree in mechatronical engineering and the B.E. degree in mechanical engineering in 2020, 2014 from Tsinghua University, Beijing, China. His research interests include three-dimensional computer vision and robot learning.
\end{IEEEbiography}
\vspace{-12mm}
\begin{IEEEbiography}[{\includegraphics[width=1in,height=1.25in,clip,keepaspectratio]{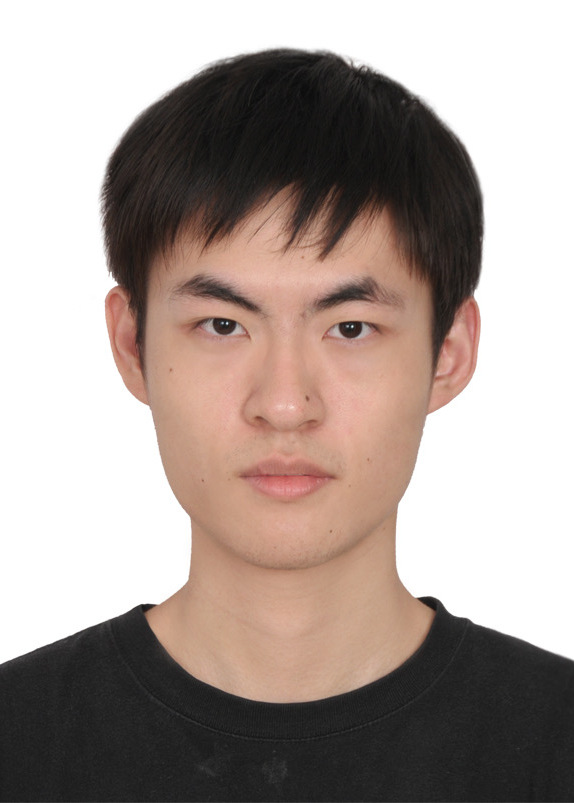}}]{Ang Li} 
 is an undergraduate student in the Department of Computer Science and Engineering at UC San Diego. His research interests lie in the intersection of computer vision, computer graphics, and robotics.
\end{IEEEbiography}
\vspace{-12mm}
\begin{IEEEbiography}[{\includegraphics[width=1in,height=1.25in,clip,keepaspectratio]{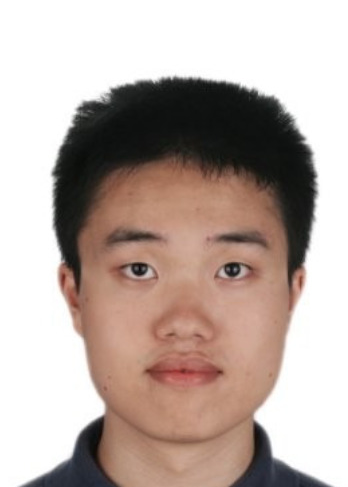}}]{Fanbo Xiang} 
received BS in computer science and BS in mathematics from University of Illinois, Urbana-Champaign. He received MS and is currently working towards a PhD degree in computer science from University of California San Diego. His research focuses on physical simulation and embodied AI platforms.
\end{IEEEbiography}
\vspace{-12mm}
\begin{IEEEbiography}[{\includegraphics[width=1in,height=1.25in,clip,keepaspectratio]{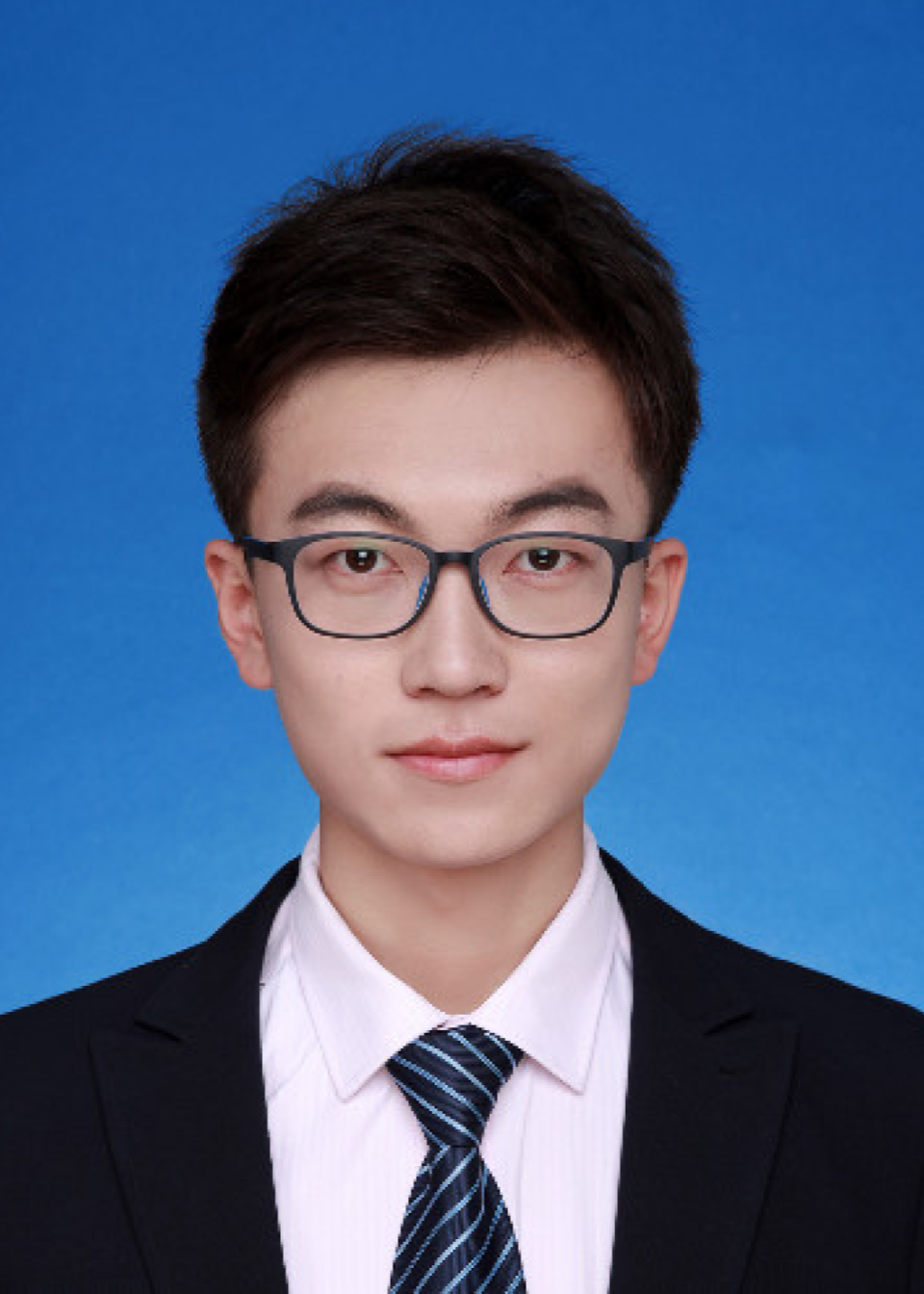}}]{Yuzhe Qin} 
received the BE degree in mechanical engineering from Shanghai Jiao Tong University, China, in 2018. He is currently working toward the PhD degree in the Department of Electrical and Computer Engineering from UC San Diego, United States. His research interests includes robot learning for manipulation, reinforcement learning with visual input, and robot teleoperation.
\end{IEEEbiography}
\vspace{-12mm}
\begin{IEEEbiography}[{\includegraphics[width=1in,height=1.25in,clip,keepaspectratio]{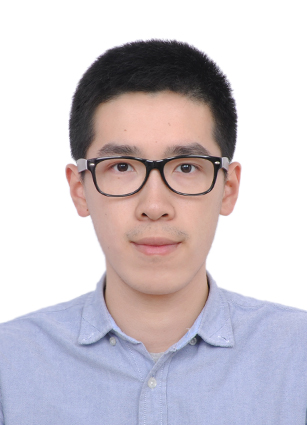}}]{Jiayuan Gu} 
received the Bachelor degree at Peking University in 2018. He is currently pursuing his Ph.D. degree in the Department of Computer Science and Engineering at the University of California, San Diego. His research interests focus on interfacing perception and policy learning algorithms for Embodied AI.
\end{IEEEbiography}
\vspace{-12mm}
\begin{IEEEbiography}[{\includegraphics[width=1in,height=1.25in,clip,keepaspectratio]{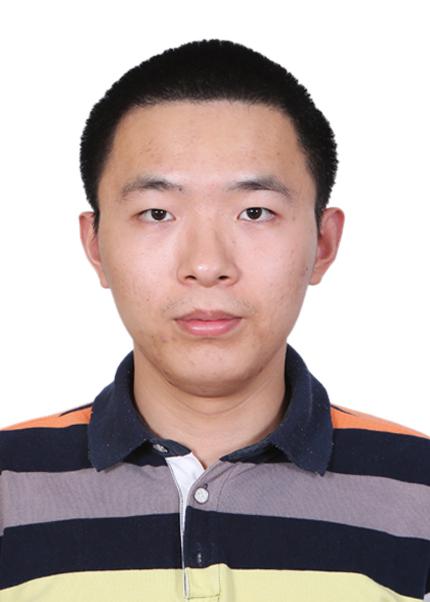}}]{Zhan Ling} 
received a B.E. degree from Institute for Interdisciplinary Information Sciences at Tsinghua University, Beijing, China, in 2019. He is currently pursuing a Ph.D. in Computer Science at the University of California, San Diego. His research interests include deep reinforcement learning and robot manipulation.
\end{IEEEbiography}
\vspace{-12mm}

\begin{IEEEbiography}[{\includegraphics[width=1in,height=1.25in,clip,keepaspectratio]{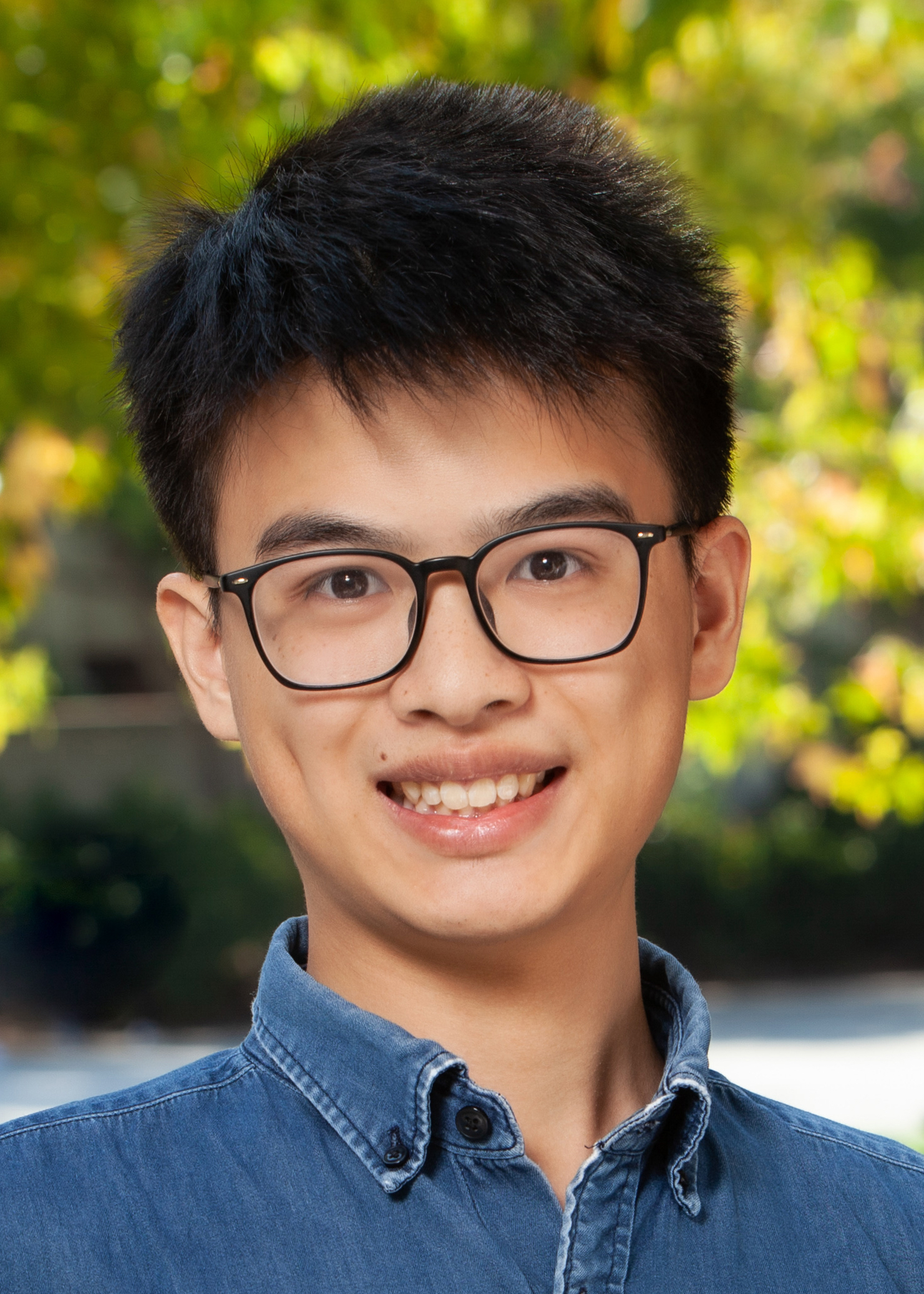}}]{Minghua Liu} 
is a PhD student in the Department of Computer Science at UC San Diego. Before that, he received his bachelor's degree in computer science from Tsinghua University. His research interests include 3D computer vision and embodied AI.
\end{IEEEbiography}
\vspace{-12mm}
\begin{IEEEbiography}[{\includegraphics[width=1in,height=1.25in,clip,keepaspectratio]{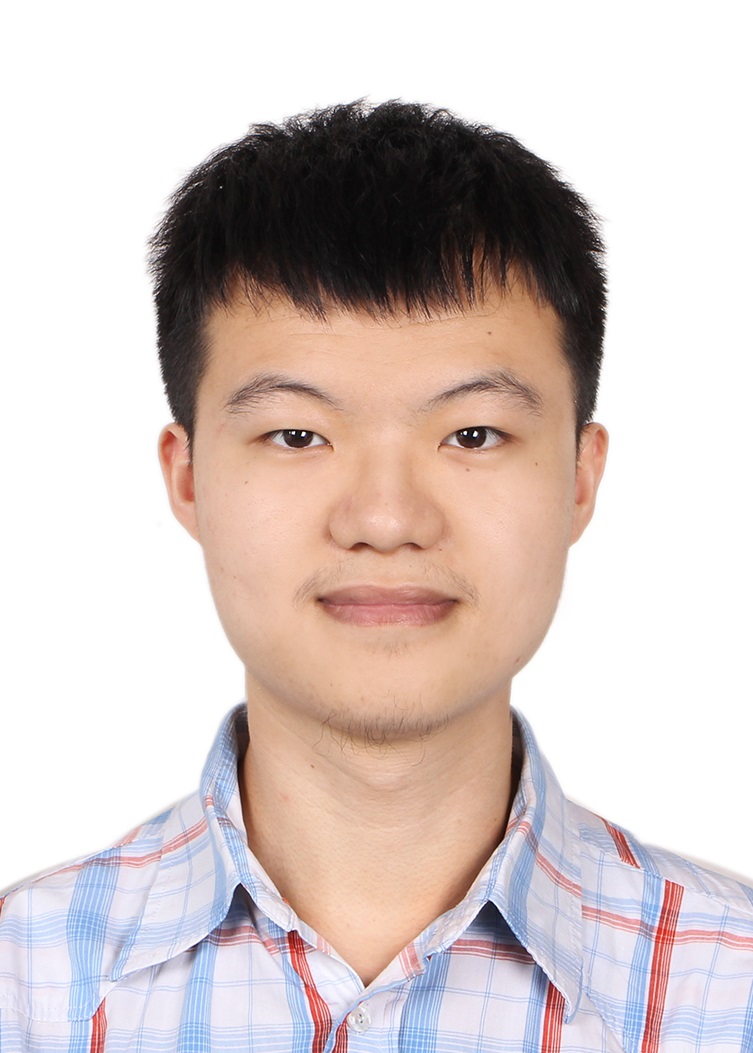}}]{Peiyu Zeng} 
received his Bachelor of Engineering in Measurement, Control Technology and Instruments from Tsinghua University, Beijing, China, in 2021. He is currently pursuing the Master of Science in Robotics, Systems and Control at ETH Zurich, Switzerland. His current research interests include model-based planning and control.
\end{IEEEbiography}
\vspace{-12mm}
\begin{IEEEbiography}[{\includegraphics[width=1in,height=1.25in,clip,keepaspectratio]{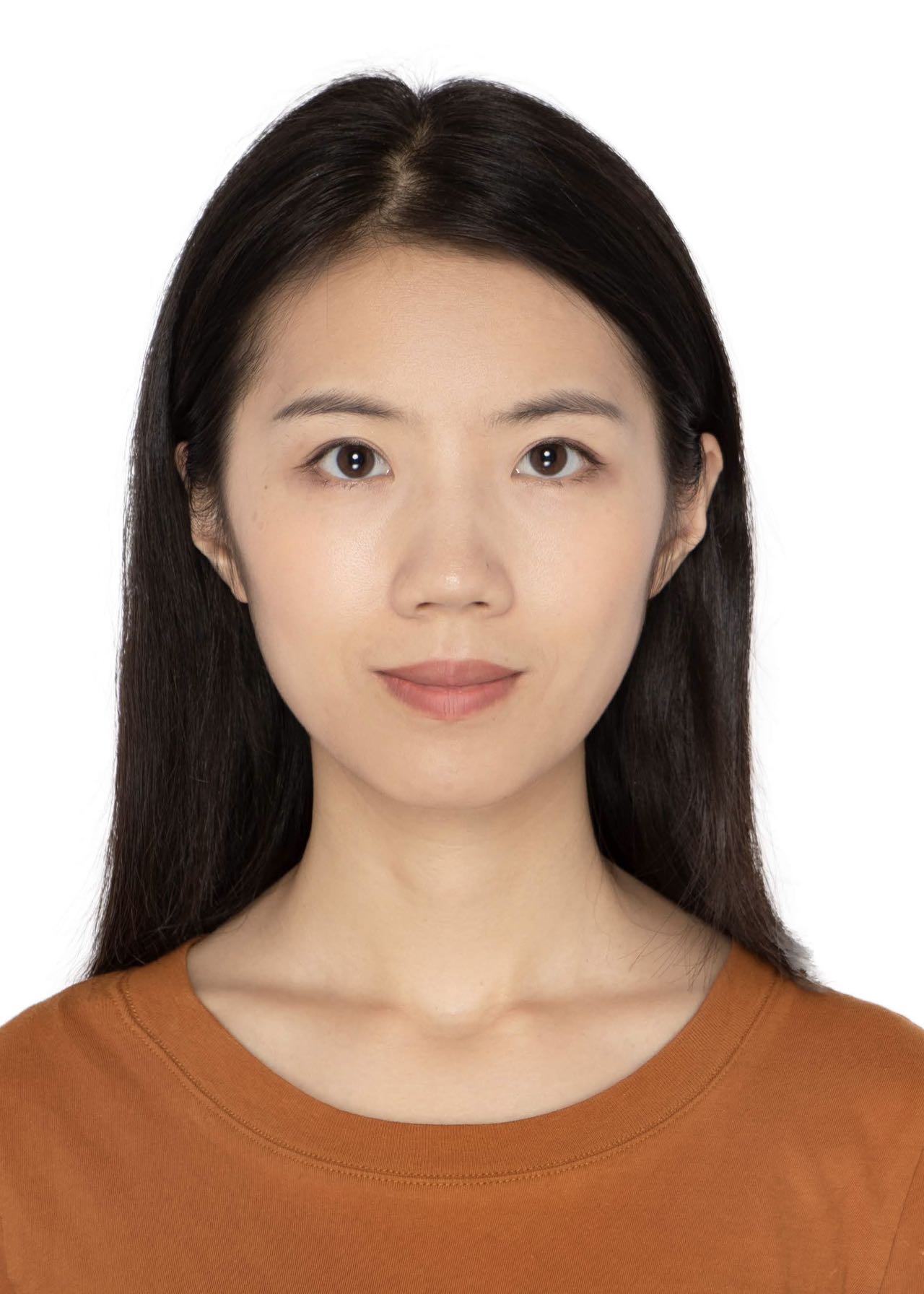}}]{Songfang Han} 
received the PhD degree from HKUST. She was a postdoctoral researcher in UCSD and now is a Machine Learning Engineer at Snap Inc. Her research interests include 3D reconstruction and geometry processing.
\end{IEEEbiography}
\vspace{-12mm}
\begin{IEEEbiography}[{\includegraphics[width=1in,height=1.25in,clip,keepaspectratio]{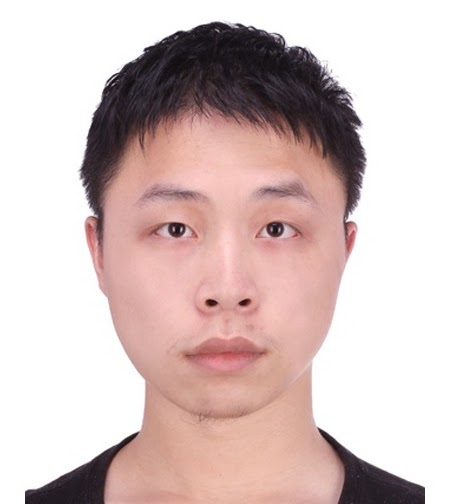}}]{Zhiao Huang} 
received his bachelor's degree at Tsinghua University, Beijing, China, in 2018. He is a currently fifth-year Ph.D. student in Computer Science at UCSD, advised by Prof. Hao Su. His research interests include soft-body manipulation with differentiable physics and reinforcement learning.
\end{IEEEbiography}
\vspace{-12mm}
\begin{IEEEbiography}[{\includegraphics[width=1in,height=1.25in,clip,keepaspectratio]{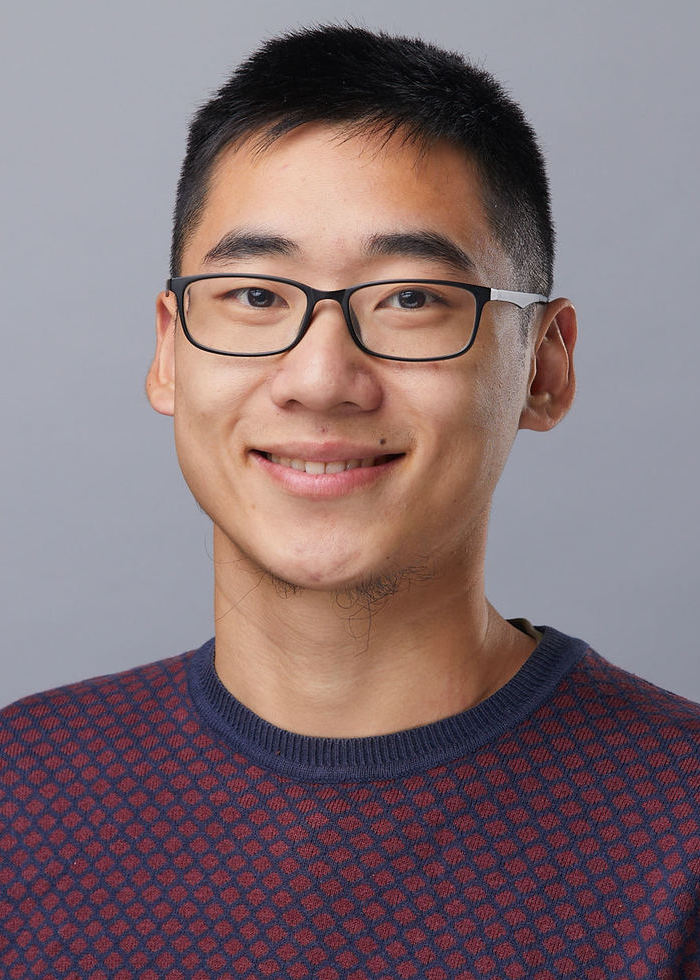}}]{Tongzhou Mu} 
 is a Ph.D. student in the Department of Computer Science and Engineering at the University of California San Diego, where he is advised by Prof. Hao Su. His research interests include reinforcement learning / imitation learning, concept discovery and reasoning, and robotics / embodied AI.
\end{IEEEbiography}
\vspace{-12mm}
\begin{IEEEbiography}[{\includegraphics[width=1in,height=1.25in,clip,keepaspectratio]{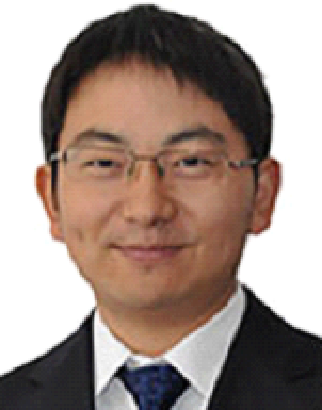}}]{Jing Xu}
received his Ph.D. in mechanical engineering from Tsinghua University, Beijing, China, in 2008. He was a Postdoctoral Researcher in the Department of Electrical and Computer Engineering, Michigan State University, East Lansing. He is currently an Associate Professor in the Department of Mechanical Engineering, Tsinghua university, Beijing, China. His research interests include vision-guided manufacturing, image processing, and intelligent robotics.
\end{IEEEbiography}
\vspace{-12mm}
\begin{IEEEbiography}[{\includegraphics[width=1in,height=1.25in,clip,keepaspectratio]{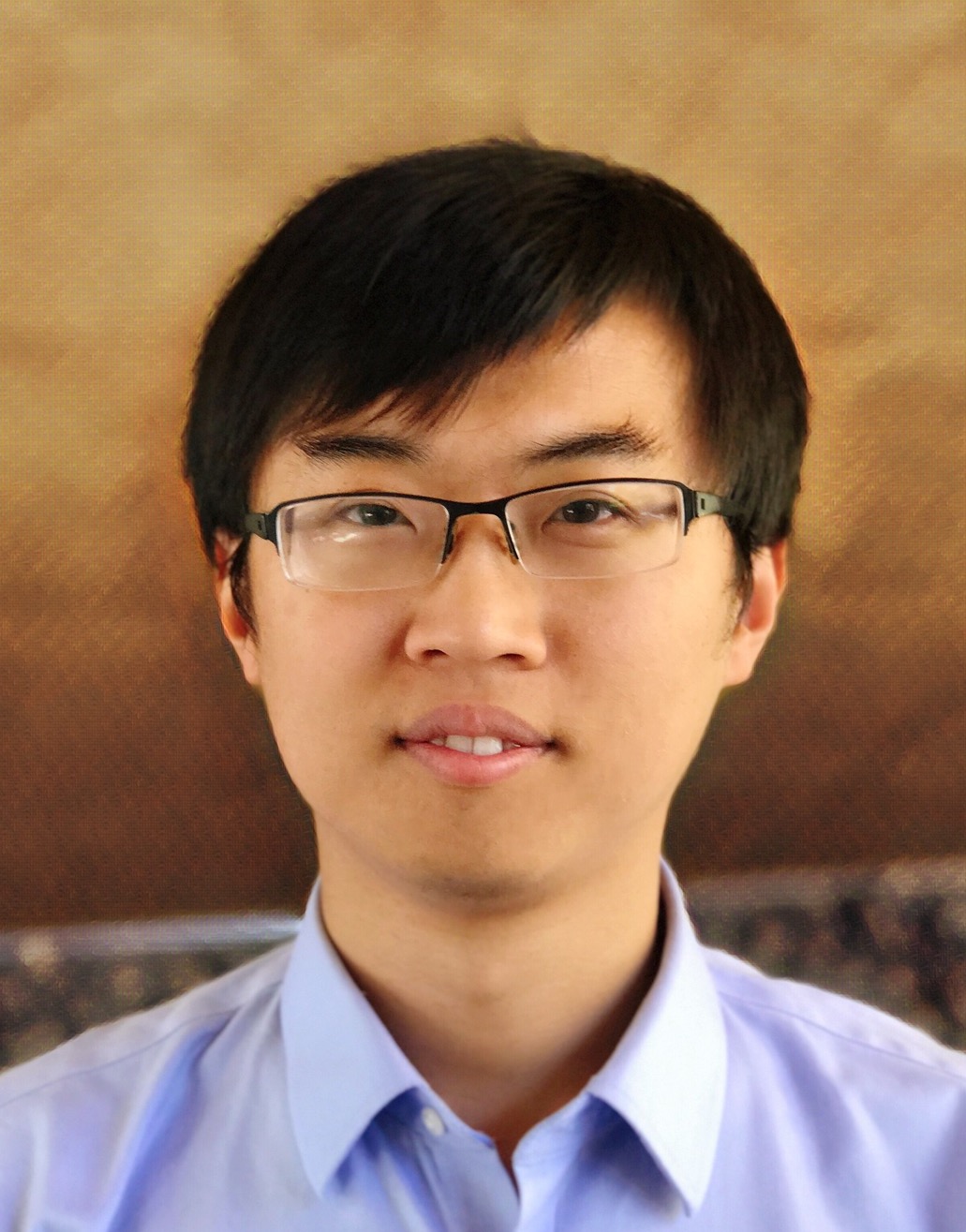}}]{Hao Su}
has been in UC San Diego as Assistant Professor of Computer Science and Engineering since July 2017. He is affiliated with the Contextual Robotics Institute and Center for Visual Computing. He served on the program committee of multiple conferences and workshops on computer vision, computer graphics, and machine learning. He is the Area Chair of ICCV'19, CVPR'19, Senior Program Chair of AAAI'19, IPC of Pacific Graphics'18, Program Chair of 3DV'17, Publication Chair of 3DV'16, and chair of various workshops at CVPR, ECCV, and ICCV. He is also invited as keynote speakers at workshops and tutorials in NIPS, 3DV, CVPR, RSS, ICRA, S3PM, etc.
\end{IEEEbiography}

\vfill

\end{document}